\documentclass[journal]{IEEEtran}
\ifCLASSINFOpdf
\else
   \usepackage[dvips]{graphicx}
\fi

\usepackage{graphicx}
\usepackage{float}
\usepackage{enumitem}
\usepackage{multirow,xcolor}
\usepackage{booktabs}
\usepackage{threeparttable}
\usepackage{color}
\usepackage{multirow}
\usepackage{algorithm}
\usepackage{algpseudocode}
\usepackage{amsmath}

\makeatletter
\def\ps@IEEEtitlepagestyle{
  \def\@oddfoot{\mycopyrightnotice}
  \def\@evenfoot{}
}
\def\mycopyrightnotice{
  {\footnotesize
  \begin{minipage}{\textwidth}
  \centering
  Copyright~\copyright~20xx IEEE. Personal use of this material is permitted. However, permission to use this  \\ 
  material for any other purposes must be obtained from the IEEE by sending a request to pubs-permissions@ieee.org.
  \end{minipage}
  }
}

\begin{document}

\title{Exploiting Global Camera Network Constraints for Unsupervised Video Person Re-identification}

\author{Xueping Wang, Rameswar Panda, Min Liu, Yaonan Wang and Amit K. Roy-Chowdhury, \IEEEmembership{Fellow, IEEE}
\thanks{Xueping Wang, Min Liu and Yaonan Wang are with the College of Electrical and Information Engineering at Hunan University and National Engineering Laboratory for Robot Visual Perception and Control Technology, Changsha, Hunan, China. ({\it{Corresponding author: Min Liu}})

Amit K. Roy-Chowdhury and Rameswar Panda are with the Department of Electrical and Computer Engineering at the University of California, Riverside.

E-mails: (wang\_xueping@hnu.edu.cn, rpand002@ucr.edu, liu\_min@hnu.edu.cn, yaonan@hnu.edu.cn, amitrc@ece.ucr.edu)
}}

\maketitle

\begin{abstract}
Many unsupervised approaches have been proposed recently for the video-based re-identification problem since annotations of samples across cameras are time-consuming. However, higher-order relationships across the entire camera network are ignored by these methods, leading to contradictory outputs when matching results from different camera pairs are combined. In this paper, we address the problem of unsupervised video-based re-identification by proposing a consistent cross-view matching (CCM) framework, in which global camera network constraints are exploited to guarantee the matched pairs are with consistency.
Specifically, we first propose to utilize the first neighbor of each sample to discover relations among samples and find the groups in each camera. Additionally, a cross-view matching strategy followed by global camera network constraints
is proposed to explore the matching relationships across the entire camera network. Finally, we learn metric models for camera pairs progressively by alternatively mining consistent cross-view matching pairs and updating metric models using these obtained matches. Rigorous experiments on two widely-used benchmarks for video re-identification demonstrate the superiority of the proposed method over current state-of-the-art unsupervised methods; for example, on the MARS dataset, our method achieves an improvement of 4.2\% over unsupervised methods, and even 2.5\% over one-shot supervision-based methods for rank-1 accuracy.
\end{abstract}


\IEEEpeerreviewmaketitle

\section{Introduction}
\IEEEPARstart{P}{erson} re-identification (re-id) is a cross-camera instance retrieval problem which aims at searching persons across multiple cameras \cite{roy2012camera,arxiv20reidsurvey}. 
In recent years, video-based person re-id has attracted increasing attention because video data provides richer information than images and it is easier to obtain than before.
Some video-based person re-id methods have been proposed and achieved impressive results \cite{chen2019spatial, rao2019learning, ye2018race,ye2019dynamic,li2019global,ouyang2018video,ye2020purifynet, chen2018bmvc,li2020hierarchical,ye2017dynamic, riachy2019video, wu2020tracklet,tip20mace,wang2020learning,pami20embedding}, however, their performance largely depends on huge amount of labeled data which are difficult to collect in real world applications. Consequently, in this work, we aim to develop a fully unsupervised solution for video person re-id that does not require any identity labels. 

\begin{figure}
\centerline{\includegraphics[width=\columnwidth]{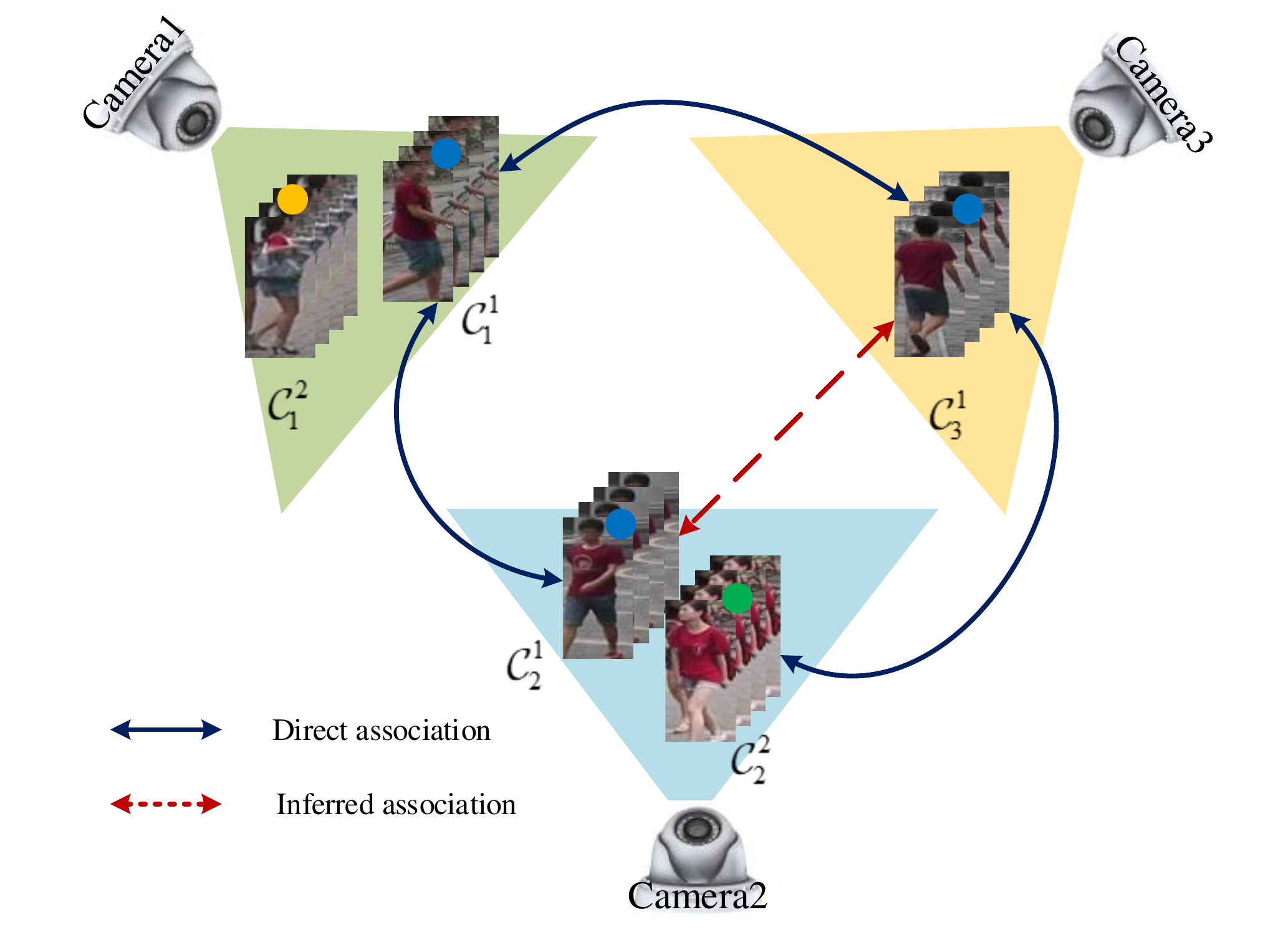}}
\caption{An illustrative example of the contradictory matches in a camera network. Different dots indicate the identity associations. $\mathcal C_r^i$ denotes sample {\em i} captured in camera {\em r}. Assuming that the cross-camera positive matching associations  ($\mathcal C_1^1,\mathcal C_2^1$), ($\mathcal C_1^1,\mathcal C_3^1$) and ($\mathcal C_2^2,\mathcal C_3^1$) can be obtained independently by using some label estimation methods shown in blue lines. We can infer that ($\mathcal C_2^1,\mathcal C_3^1$) is also a positive match because they are matched to the same sample $\mathcal C_1^1$  shown in red line. However when combining them together, there is an infeasible scenario that indicates that $\mathcal C_2^1$ and $\mathcal C_2^2$ are with the same label. Best viewed in color.}
\label{fig:contradictory matches}
\end{figure}

Recently, some unsupervised person re-id methods have been proposed that exploit unlabeled data progressively by assigning pseudo-labels and updating the re-id model in an alternative manner
\cite{lin2019bottom,fan2018unsupervised,yu2019unsupervised}.
Despite promising results on common benchmarks, most of these methods are not fully unsupervised and still require some label information, such as source domain labeled data (domain adaption-based unsupervised method) \cite{fu2019self, zhang2019self}, to train a model, which limits the scalability of prior methods in practical applications. 
In recent years, some cross-camera matching methods have been proposed for person re-id or object tracking in a camera network and they achieved impressive performance \cite{zhang2016prism,chu2014fully,lin2020unsupervised}. However, most of these methods only consider the intra-camera and inter-camera matching correlations of samples independently \cite{zhang2016prism,chu2014fully,lin2020unsupervised}, but ignore the higher-order relationships across the entire camera network. 
This may lead to contradictory outputs when matching results from different camera pairs are combined.

To illustrate this further, consider Figure \ref{fig:contradictory matches} which shows a camera network containing 3 cameras and each of them capture 1-2 persons. Assume that the cross-camera positive matching associations between $(\mathcal C_1^1,\mathcal C_2^1)$, $(\mathcal C_2^2,\mathcal C_3^1)$ and $(\mathcal C_1^1,\mathcal C_3^1)$ can be obtained independently by using some label estimation methods ($\mathcal C_r^i$ denotes $i$th person captured in camera {\em r}). We can infer that ($\mathcal C_2^1,\mathcal C_3^1$) is also a positive matching pair because they are matched to the same person $\mathcal C_1^1$. However, when these matches from different camera pairs are combined, it leads to an infeasible scenario - $\mathcal C_2^1$ and $\mathcal C_2^2$ are with the same label. It is hard to distinguish which matches are reliable. Few recent methods \cite{lin2017consistent,das2014consistent,chakraborty2015network} introduce global camera network constraints into person re-id task for reducing the unreliable matches by exploring high-order relationships in a camera network. However, they require a large number of labeled samples to train their models or the complex optimization method. Motivated by this, we ask an important question in this paper: \textit{Can we develop a reliable cross-camera label estimation strategy, in which the matches are with a guarantee of consistency, for improving the performance of unsupervised re-id without requiring any labeled samples?} This is an especially important problem in the context of many person re-id systems involving large number of cameras.

To address such problems, in this paper, we propose a consistent cross-view matching framework by exploiting global camera network constraints for unsupervised video person re-id.
First, the proposed method is fully unsupervised. We propose to use a first neighbor-based clustering strategy \cite{sarfraz2019efficient} to discover the intra-camera label relations and then cross-view matching to explore the inter-camera correlations without requiring any labeled samples for model learning. Second, our approach generates cross-view matches with a guarantee of consistency. 
Specifically, global camera network constraints are introduced into the cross-view matches to obtain the reliable matching pairs, including a definition for the reliability of matches to reduce the false positive ones. 
Finally, we learn metric models for camera pairs progressively by using an iterative updating framework which iterates between consistent cross-view matching and metric models learning. 

To summarize, the contributions of our work are as follows. 
\begin{itemize}[leftmargin=*]
    \item We propose a fully unsupervised, consistent cross-view matching framework, 
    for video person re-id, in which the estimated cross-camera positive matching pairs follow the notion of consistency.
    
    \item We propose a definition for reliability of the cross-view matches via introducing global network constraints, which can reduce the incorrect matches significantly. 
    
    \item Extensive experiments demonstrate that our approach outperforms the state-of-the-art unsupervised methods on MARS and DukeMTMC-VideoReID datasets and is very competitive while comparing with one-shot supervision-based methods. 
\end{itemize}

Note that the cross-camera label estimation task usually suffers from large inter-camera variations due to different camera environments and self appearance changes of persons. Thus, we focus on unsupervised video re-id task as video tracklets contain much richer information than images, which helps to disambiguate difficult cases that arise when trying to recognise a person in a different camera. Different from other re-id methods, our method focuses on optimizing the cross-camera matching relations in a camera network with an unsupervised consistent cross-view matching framework. By introducing the global camera network constraints, the obtained matches will be more reliable than that considering inter-camera relations independently. This strategy can be used to other cross-view label estimation tasks, such as cross-camera person re-id and cross-view object tracking \cite{zhang2016prism,chu2014fully,lin2020unsupervised}.

\begin{figure*}
\centering
\centerline{\includegraphics[width=1.7\columnwidth]{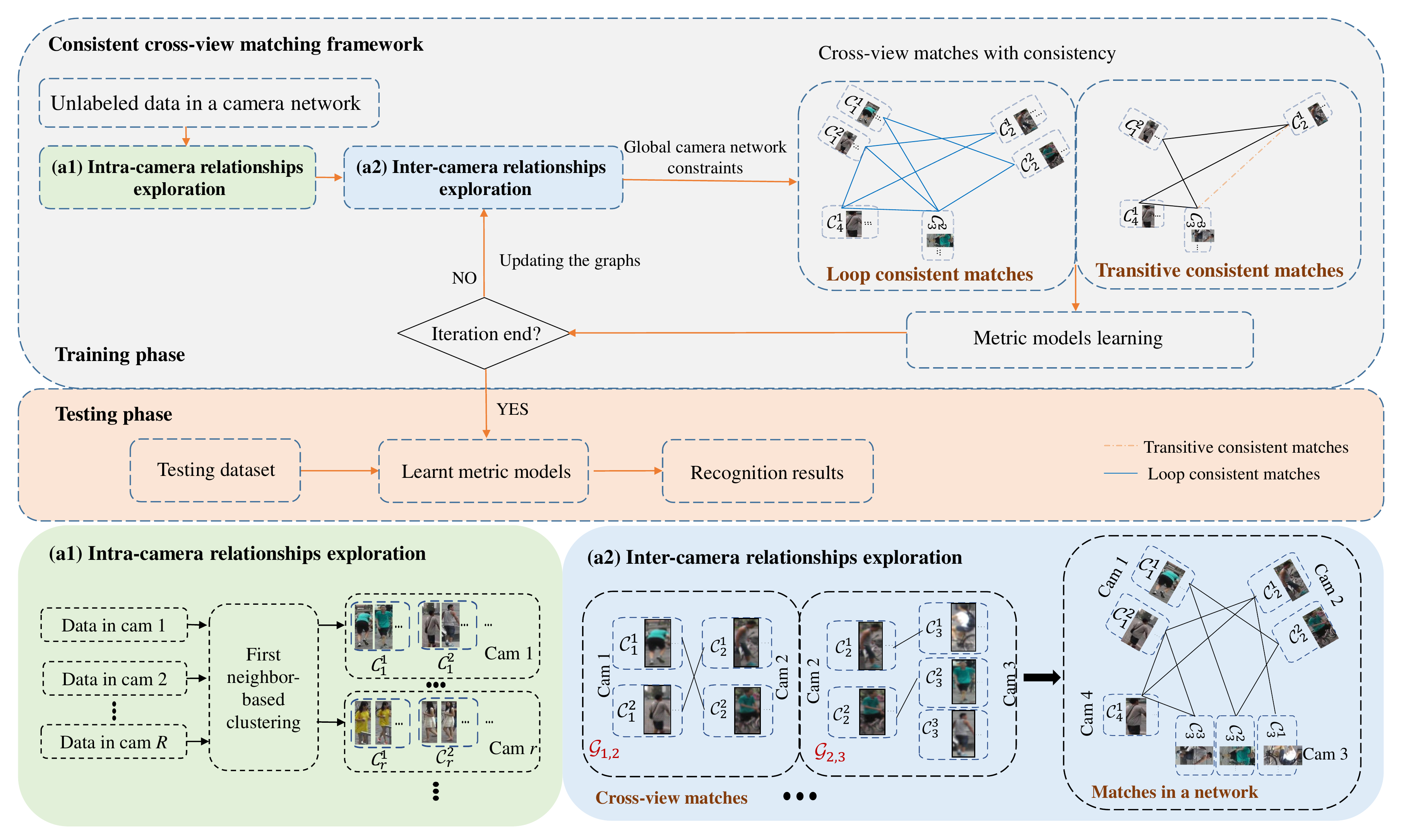}}
\caption{Overview of our proposed method. This figure demonstrates the overall framework of the proposed approach. By introducing global camera network constraints into the matches in a camera network, we can select some reliable pairs with a guarantee of consistency. Thereafter, we learn metric models for camera pairs progressively by alternatively  mining consistent cross-view matches and updating metric models. (a1) shows the intra-camera clustering for each camera. By using the first neighbor-based clustering algorithm, first neighbor relations can be obtained in each camera. According to Equation 1, the adjacency matrix can be obtained. Thereafter, the connected samples can be clustered together. $\mathcal C_r^i$ denotes $i$th cluster in camera {\em r}. (a2) illustrates the inter-camera relationships exploration across a camera network. There may be contradictory matches when combining all cross-view matches together, so we introduce global camera network constraints into these matches to obtain reliable pairs. Note that each image in this figure denotes one person tracklet. } 
\label{fig:framework}
\end{figure*}

\section{Related Work}
\subsection{Unsupervised  Person Re-id}
Unsupervised learning models have recently received much attention in person re-id task as they do not require manually labeled data.  
Most of the proposed unsupervised person re-id methods exploit  unlabeled  data  progressively by assigning  pseudo-labels and updating re-id  models in an alternative manner.
Fan et al. \cite{fan2018unsupervised} proposed a {\em k}-means clustering-based method to select reliable images gradually and use them to fine tune a deep neural network to learn discriminative features for person re-id. Lin et al. \cite{lin2019bottom,liny2020unsupervised} proposed a hierarchical clustering-based feature embedding method by regarding sample labels as supervision signals to train a non-parametric convolutional neural network \cite{xiao2017joint}. Liu et al. \cite{liu2017stepwise} presented a person re-id method which iterates between cross-camera tracklet association and feature learning. 
Li et al. \cite{li2018unsupervised} proposed a deep learning based tracklet association method by jointly learning per-camera tracklet association and cross-camera tracklet correlation to obtain the label information. 

Some  cross-camera  matching  methods  have been proposed for person re-id and they obtained impressive performance. Lin et al. \cite{lin2020unsupervised} proposed a cross-camera encouragement (CCE) term to assign different distances to image pairs from different cameras, which explores the cross-camera relations and overcomes the negative effect caused by the wrong clustering results. In \cite{liny2020unsupervised}, images from different camera styles are generated for data augmentation, so that the relations between cameras are inferred during unsupervised training.

Generative adversarial networks have also been adopted to train a camera style transfer model to bridge the gap between the labeled source domain and unlabeled target domain. Zhong et al. \cite{zhong2018camera} introduced camera style adaptation as a data augmentation approach that smooths
the camera style disparities. Deng et al.  \cite{deng2018image} translated the labeled images from source to target domain in an unsupervised manner and then trained re-ID models with the translated images by supervised
methods.

Despite promising results on common benchmarks, most of these methods ignore the high-order relationships in a camera network and still require some person identity information for their models training. On the contrary, our approach introduces the global camera network constraints into person re-id models to explore more reliable cross-camera sample pairs and it is a fully unsupervised method that does not require any identity labeled data.

\subsection{Graph Matching for Person Re-id}
{Graph matching has been widely used in computer vision and machine learning domains, such as shape matching and object recognition \cite{berg2005shape,yan2015multi,zhang2016pairwise}. 
Recently, several works also introduce it into the person re-id task. 
Wu et al. proposed an unsupervised graph association method \cite{wu2019unsupervised} to mine the cross-view relationships and reduce the damage of noisy associations.
Ye et al. \cite{ye2019dynamic} presented a dynamic graph co-matching method to obtain the corresponding image pairs across cameras.
Das et al. \cite{das2014consistent,chakraborty2015network} proposed a consistent re-id method in a camera network by considering the matching consistency to improve camera pairwise re-id performance. 
Following \cite{das2014consistent}, Lin et al. \cite{lin2017consistent} proposed a consistent-aware deep learning method by incorporating consistency constraints into deep learning framework for person re-id. 
Roy et al. \cite{roy2018exploiting} constructed a {\em k}-partite graph for the camera network and then exploited transitive information across the graph to select an optimal subset of image pairs for manual labeling. 
However, most of these methods often ignored the higher-order relationships in a camera network. The proposed approach introduced the global network constraints into cross-view matches, which can reduce the incorrect matches significantly and learn robust metric models for camera pairs progressively.
}

\section{Consistent Cross-View Matching}
\label{sec:guidelines}
In this section, we first explore the intra-camera label relationships by using a first neighbor-based clustering strategy. On top of that, we construct a graph for each pair of cameras and search for reliable cross-view matching pairs with consistency by introducing global camera network constraints. Finally, we learn the distance metric models for camera pairs progressively by alternatively mining consistent matches and updating the learned metric models. The overall framework of our proposed method is shown in Figure \ref{fig:framework}. 

In camera $r$, we assume that there are $N_r$ samples and denote it as $\mathcal I_r=\left\{I_r^1,I_r^2,...,I_r^{N_r}\right\}$. A pre-trained feature embedding model $f(\cdot)$ is employed to extract features for the training samples $\mathcal T_r=\left\{T_r^1,T_r^2,...,T_r^{N_r}\right\}$ and the extracted features are used as the inputs of our approach.

\subsection{Intra-camera Relationships Exploration}
In each camera, there is not much appearance variation between the samples with the same identity. So, we propose to utilize the first neighbor of each sample which can be obtained via fast approximate nearest neighbor methods (such as k-d tree) to explore the label relationships among samples and find the groups in each camera \cite{sarfraz2019efficient}. 
Specifically, given the indexes of the first neighbor of each sample in one camera, we define an adjacency matrix:

\begin{equation}
    A(i,j)=
    \left\{
    \begin{array}{lr}
    1, & \mbox{if } i=k_j^1 \mbox{ or } j=k_i^1 \mbox{ or } k_i^1=k_j^1;\\
    0, &\mbox{otherwise}.
    \end{array}
    \right.
\end{equation}
where $k_j^1$ denotes that the first neighbor of sample $j$. The adjacency matrix links each sample {\em i} to its first neighbor via $j = k_i^1$ , enforces symmetry via $k_j^1= i$ and links samples $(i; j)$ that have the same neighbor with $k_i^1=k_j^1$. 
Equation 1 for each camera returns a symmetric sparse matrix directly specifying a graph with connected components as the clusters (shown in Figure \ref{fig:framework} (a1)). It is reasonable to regard each cluster as one person.
So, one camera, e.g. camera {\em r}, can be denoted as $\mathcal C_r=\{\mathcal C_r^1,\mathcal C_r^2,...,\mathcal C_r^{n_r}\}$ with $n_r$ clusters/persons, where $\mathcal C_r^i$ is the $i$th cluster/person in camera {\em r}.

\subsection{Inter-camera Relationships Exploration}

\subsubsection{Graph Construction} We construct a bipartite graph $\mathcal G=(U,V,E)$ for each pair of cameras where each part of the graph denotes one camera and the vertices are the obtained clusters/persons (in section A). For example, we  could convert camera pair $(p,q)$ into a graph $\mathcal G_{p,q}=(\mathcal C_p,\mathcal C_q,E_{M_{p,q}})$, where $\mathcal C_p=\{\mathcal C_p^1,\mathcal C_p^2,...,\mathcal C_p^{n_p}\}$ and $\mathcal C_q=\{\mathcal C_q^1,\mathcal C_q^2,...,\mathcal C_q^{n_q}\}$ denote camera $p$ and $q$, respectively. 
Note that we will use the terms `cluster', ‘person’ and ‘vertex’ interchangeably throughout our work. 
The edge $E_{M_{p,q}}$ is a matching cost matrix of camera pair $(p,q)$ and each element $e_{M_{p,q}}^{i,j}$ describes the similarity of vertex pair $(\mathcal C_p^i,\mathcal C_q^j)$ which is computed through a minimum distance criterion that takes the shortest distance between samples in two clusters, as follows:
\begin{equation}
    e_{M_{p,q}}^{i,j}=\mathop {\min }\limits_{\mathop {T_p^a}\in\mathcal C_p^i,\mathop {T_q^b}\in\mathcal C_q^j} d_{M_{p,q}}(T_p^a,T_q^b)
\end{equation}
where $M_{p,q}$ denotes a distance metric model learned using the estimated pairs with consistency from camera $p$ and $q$, which is initialized with identity matrix, and $d_{M_{p,q}}(T_p^a,T_q^b)=(T_p^a-T_q^b)^TM_{p,q}(T_p^a-T_q^b)$.

\subsubsection{Graph matching}
{We use the assignment matrix $X_{p,q}$ to represent the matching associations between the vertices across camera pair $(p,q)$.
Element $x_{p,q}^{i,j}$ in $X_{p,q}$ represents the matching association of the vertex $\mathcal C_p^i$ and $\mathcal C_q^j$, which is defined as follows:
\begin{equation}
    x_{p,q}^{i,j}=
    \left\{
    \begin{array}{lr}
    1, & \mbox{if } \mathcal C_p^i \mbox{ and } \mathcal C_q^j \mbox{ are a matched pair};\\
    0, &\mbox{otherwise}.
    \end{array}
    \right.
\end{equation}
}

In a large camera network, it is common that one camera may not capture every person. In this situation, a person from one camera {\em p} can have at most one match from another camera {\em q}. In other words, the matching association values in every row or column of the assignment matrix $X_{p,q}$ can all be 0. As a result, the matching association constraints are as follows:
\begin{equation}
    {\sum_{j=1}^{n_q}x_{p,q}^{i,j}\leq 1,i=1,2,...,n_p} \mbox{ and }  \\
    {\sum_{i=1}^{n_p}x_{p,q}^{i,j}\leq 1,j=1,2,...,n_q}\\
\end{equation}
where $n_p$ and $n_q$ are the number of persons/clusters in camera {\em p} and camera {\em q}, respectively.

To compute the assignment matrix across camera pairs, we formulate it as a binary linear programming with constraints as follows:
\begin{equation}
\begin{array}{l}
X_{p,q}=\mathop {\arg \min }\limits_{\mathop {x_{p,q}^{i,j}}}  {\sum\limits_{i,j = 1}^{n_p,n_q} {e_{M_{p,q}}^{i,j}x_{p,q}^{i,j}} } \\
{\mbox {subject to:  }}x_{p,q}^{i,j} \in \{ 0,1\} ,\forall i = 1,...,n_p,j=1,...,n_q\\
\mbox{\qquad \qquad       }\sum\limits_{i = 1}^{n_p} {x_{p,q}^{i,j} \leq 1,} \forall j = 1,...,n_q\\
{\rm{  \qquad \qquad   }}\sum\limits_{j = 1}^{n_q} {x_{p,q}^{i,j} \leq 1,} \forall i = 1,...,n_p\\
\end{array}
\end{equation}
The assignment matrix set $\mathbf{X}=\{X_{p,q}|p<q\}$ across the pair of cameras in a network can be obtained, where  $X_{p,q}=\{x_{p,q}^{i,j}|i=1,...,n_p,j=1,...,n_q\}$.

\subsection{Global Camera Network Constraints}
Existing methods, like Hungarian algorithm \cite{kuhn1955hungarian} can be directly used to solve the above binary linear programming problem.
However, it is hard to ensure that the obtained matching associations are reliable because Hungarian algorithm will try to get as many matching associations as possible. Thus, the assignment matrix may contain a lot of false positive matches. In addition, these cross-view matched pairs also do not consider matching consistency in a network of camera. 
It may lead to contradictory outputs when matching associations from different camera pairs are combined as shown in Figure \ref{fig:contradictory matches}. 
To address this problem, we introduce global camera network constraints including loop consistency constraints and transitive inference consistency constraints into these cross-view matches, which will guarantee the obtained cross-view matching pairs are with consistency. 

\subsubsection{Loop consistent matches}
\begin{figure}
\centerline{\includegraphics[width=\columnwidth]{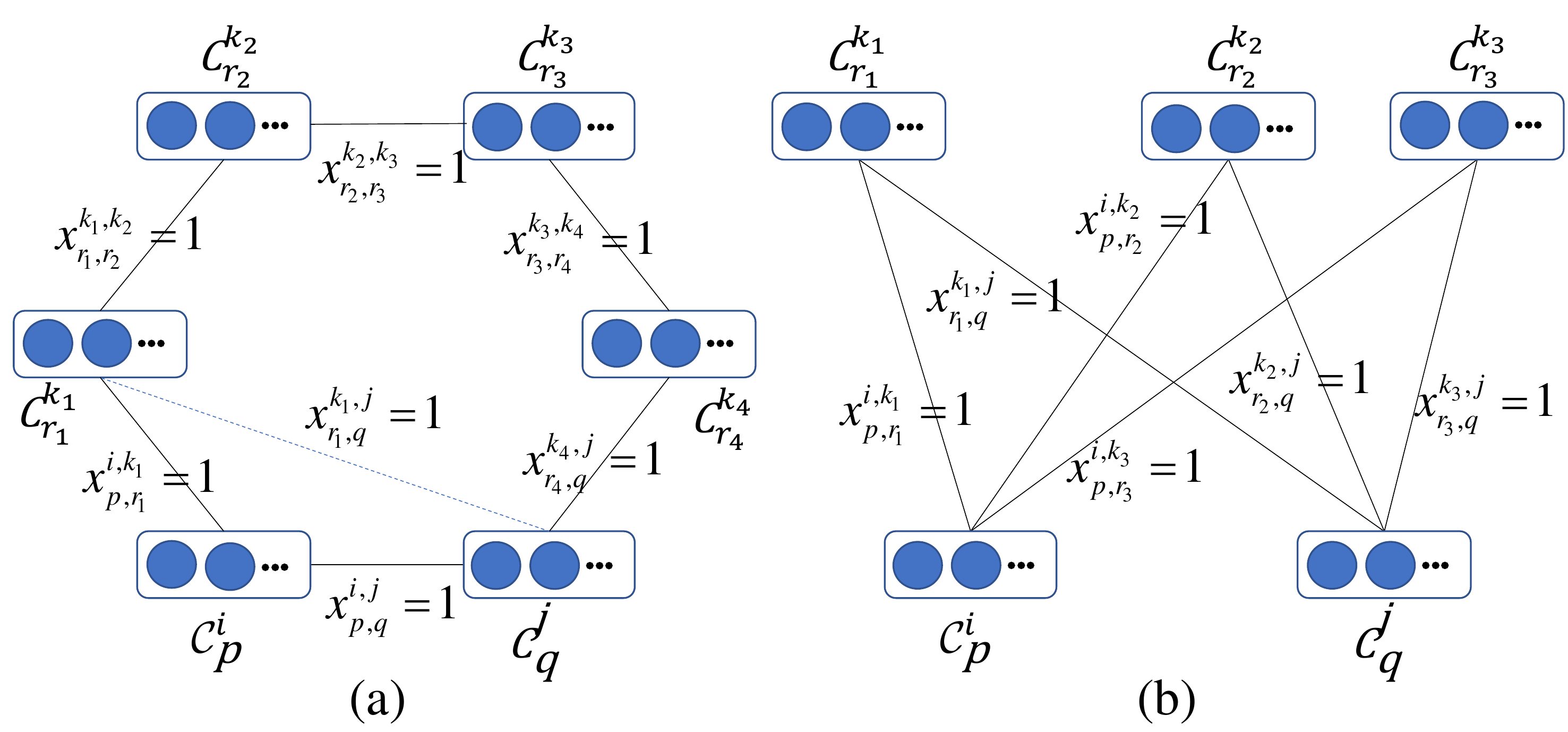}}
\caption{An example of consistent cross-view matches. (a) demonstrates loop consistent constraint. If $x_{p,q}^{i,j}=1$ and existing a person $k_1$ in camera $r_1$ satisfies $x_{p,r_1}^{i,k_1}x_{r_1,q}^{k_1,j}=1$, the match $(\mathcal C_p^i,\mathcal C_q^j)$ is with consistency. (b) shows a transitive inference consistency pair $(\mathcal C_p^i,\mathcal C_q^j)$ and $\mbox{RT}_{p,q}^{i,j}=3$.}
\label{fig:consistency}
\end{figure}

{Given two vertices $\mathcal C_p^i$  and $\mathcal C_q^j$ from a camera pair $(p,q)$ in a camera network, it can be noted that for consistency, logical `AND' relationship between the association value $x_{p,q}^{i,j}$ and the set of association values $\{x_{p,r_1}^{i,k_1},x_{r_1,r_2}^{k_1,k_2},...,x_{r_n,q}^{k_n,j}\}$ across possible vertices in different cameras has to be maintained, where $r_1,...,r_n,p,q$ denote cameras in a network and $k_1,...,k_n,i,j$ represent the persons captured by corresponding cameras.
In other words, the association value $x_{p,q}^{i,j}$ between the two vertices $\mathcal C_p^i$ and $\mathcal C_q^j$ has to be 1, and it has to satisfy the indirect matching association $x_{p,r_1}^{i,k_1}x_{r_1,r_2}^{k_1,k_2}...x_{r_4,q}^{k_4,j}=1$ as shown in Figure \ref{fig:consistency} (a). 
In \cite{das2014consistent,chakraborty2015network}, it has been proven that if the loop consistency constraint is satisfied for every triplet of cameras, it automatically ensures consistency for every possible combination of cameras taking 3 or more of them. Thus, the consistent matching pair $\mathcal C_p^i$ and $\mathcal C_q^j$ in the network of cameras has to satisfy the direct cross-view matching association $x_{p,q}^{i,j}=1$ and a person {\em k} in camera {\em r} should satisfy: $x_{p,r}^{i,k}x_{r,q}^{k,j}=1$ as shown in Figure \ref{fig:consistency}(a). 
\begin{equation}
    x_{p,q}^{i,j}=1 \mbox{ and } \exists \mathcal{C}_r^k, x_{p,r}^{i,k}x_{r,q}^{k,j}=1,r\neq p,q
\end{equation}
}

\subsubsection{Transitive inference consistent matching}
Transitive inference among person identities across multiple cameras and their logical consequences are strongly informative properties \cite{roy2018exploiting}. We exploit the transitive relations for enhancing the performance of our cross-view matches. To illustrate the idea, let us consider a plausible scenario as shown in Figure \ref{fig:consistency}(b). Assuming we have positive cross-view matches 
$(\mathcal C_p^i,\mathcal C_{r_1}^{k_1})$ and $(\mathcal C_{r_1}^{k_1},\mathcal C_q^j)$, then according to the transitive inference we can directly infer that $\mathcal C_p^i$ and $\mathcal C_q^j$ also have the same label,
i.e., $x_{p,r_1}^{i,k_1}x_{r_1,q}^{k_1,j}=1 \Rightarrow x_{p,q}^{i,j}=1$.
Obviously, by introducing transitive inference, we can increase the number of cross-view matching pairs in a camera network.
Usually, in a camera network, with more than two cameras, we define the reliability of the transitive inference-based cross-view matches
\begin{equation}
    \mathrm{RT}_{p,q}^{i,j}=\sum_{r}\sum_{k=1}^{n_r} x_{p,r}^{i,k}x_{r,q}^{k,j},p\neq q \mbox{ and } r\neq p,q
\end{equation} 
where {\em {p, q}} and {\em r} are cameras in a network. $n_r$ is the number of persons/clusters in the camera {\em r}. $\mathrm{RT}_{p,q}^{i,j}$ denotes the reliability of the pair $(\mathcal C_{p}^{i},\mathcal C_{q}^{j})$. The larger the value $\mathrm{RT}_{p,q}^{i,j}$ is, the more reliable the transitive inference-based match is, as shown in Figure \ref{fig:consistency}(b), i.e. $\mbox{RT}_{p,q}^{i,j}=3$. When the reliability value $\mbox{RT}_{p,q}^{i,j}$ satisfies $\mbox{RT}_{p,q}^{i,j}>1$, we regard the matching pair $(\mathcal C_{p}^{i},\mathcal C_{q}^{j})$ as a transitive inference consistent match.

Note that loop consistency can be regarded as a specific form of the transitive inference consistency constraints.
Therefore, combining them together, we define a metric for measuring the reliability of  cross-camera matched pairs as follows:
\begin{equation}
    \mathrm{RLT}_{p,q}^{i,j}=x_{p,q}^{i,j}+\sum_{r}\sum_{k=1}^{n_r} x_{p,r}^{i,k}x_{r,q}^{k,j},r\neq p,q,
\end{equation} 
where $\mbox{RLT}_{p,q}^{i,j}$ represents the reliability of the cross-view matched pair $(\mathcal C_{p}^{i},\mathcal C_{q}^{j})$. 
The larger the value is, the more reliable the match is. With this reliability score, we obtain the consistent assignment matrices 
$\hat X_{p,q}=\{\hat x_{p,q}^{i,j}|i=1,...,n_p,j=1,...,n_q\}$ to learn metric models for camera pairs as,
\begin{equation}
    \hat x_{p,q}^{i,j}=
    \left\{
    \begin{array}{lr}
    1, & \mbox{if } \mathrm {RLT}_{p,q}^{i,j}>\theta;\\
    0, &\mbox{otherwise}.
    \end{array}
    \right.
\end{equation}
where $\theta$ is a threshold that is used to balance the quality and quantity of the selected matches. Obviously, with the increase in $\theta$ value, the selected pairs will be more reliable, however, the number of the matches will be less for training.
Thus, we can obtain sufficient and reliable cross-view matches in a camera network by selecting a suitable $\theta$ value for the unsupervised video person re-id task.

\subsection{Metric Learning with Consistent Matches}
Given a consistent assignment matrix $\hat X_{p,q}$ for camera pair $(p,q)$, the corresponding metric model $M_{p,q}$ could be learned to update its matching cost matrix $E_{M_{p,q}}$. In this paper, we use the log-logistic metric learning as the loss function \cite{liao2015efficient},
\begin{equation}
    f_{M_{p,q}}(\mathcal C_p^i,\mathcal C_q^j)=\log(1+e^{\hat x_{p,q}^{i,j}
    (e_{M_{p,q}}^{i,j}-\mu_{p,q})})
\end{equation}
where $e_{M_{p,q}}^{i,j}$ is the minimum distance between clusters $\mathcal C_p^i$ and $\mathcal C_q^j$ as calculated in Equation 2. $\mu_{p,q}$ is the average distance of all consistent matches from camera pair $(p,q)$.
Accordingly, for the camera pair $(p,q)$, the overall cost function is
\begin{equation}
    F(M_{p,q};\hat X_{p,q})=
    \sum_{i=1}^{n_p}\sum_{j=1}^{n_q}w_{i,j}f_{M_{p,q}}(\mathcal C_p^i,\mathcal C_q^j),M_{p,q}\succeq 0
\end{equation}
where $w_{i,j}$ is utilized to handle the imbalanced positive and negative pairs, i.e. $w_{i,j}=\frac{1}{N_{pos}}$ if $\hat x_{p,q}^{i,j}=1$, and $\frac{1}{N_{neg}}$ otherwise, and $N_{pos}$ and $N_{neg}$ are the number of consistent matches and negative pairs.

During testing, we compute the distance of each query-gallery pair $(T_{qu},T_{ga})$ by taking the minimum value under different pair-wise distance metric models as follows:
\begin{equation}
    D(T_{qu},T_{ga})=
    \mathop{\min}\limits_{\mathop{p,q=1,...,R}\limits_{\scriptstyle {p<q} \hfill\atop}}\{d_{M_{p,q}}(T_{qu},T_{ga})\}
\end{equation}
\subsection{Iterative Updating}
{In this work, we learn metric models for camera pairs progressively by alternatively mining consistent cross-view matches and updating metric models. 
In each iteration, the learned metric models are used to update the corresponding matching cost matrix in Equation 5 for better exploring inter-camera relationships in a new iteration. Thereafter, the updated consistent cross-view matching correlations could be used to update the previous metric models. Finally, the reliable cross-view matches with consistency and distance metric models can be obtained.

\subsubsection{Convergence Analysis} Note that we have two objective functions $F$ and $G$ for optimizing $M$ and $X$ in each iteration. To ensure the overall convergence of the proposed method, we adopt the same strategy as discussed in \cite{tian2012convergence}. $M$ can be optimized by choosing a suitable working step size $\eta\leq L$, where $L$ is the Lipschitz constant of the gradient function $\bigtriangledown F(M;\hat X)$. Thus, it could ensure $F(M^t;\hat X^{t-1})\leq F(M^{t-1};\hat X^{t-1})$, a detailed proof is
shown in \cite{beck2009fast}.
For $X_{p,q}^t$, the updating procedure at iteration {\it t} is constrained by keeping updating $E_{M_{p,q}}^t$ until a better $X_{p,q}$ is obtained, which satisfies $G(X_{p,q}^t;M_{p,q}^t) \leq G(X_{p,q}^{t-1};M_{p,q}^t)$ where $G(X_{p,q};M_{p,q})=E_{M_{p,q}}X_{p,q}$, and the convergence analysis has been verified from \cite{tian2012convergence}. 
}

\subsubsection{Complexity Analysis} 
The major computational cost of our CCM comes from the intra-camera and cross-view label estimation. We assume that the number of samples in each camera is {\it n}. The intra-camera label estimation complexity is $\mathcal O(n \mbox{log} n)$ \cite{sarfraz2019efficient}.
After intra-camera clustering, we assume that each camera has {\it m} clusters, then the cross-view matching time complexity is $\mathcal O(m^3)$ \cite{kuhn1955hungarian}. Updating metric model $M$ with accelerated proximal gradient is extremely fast as illustrated in \cite{beck2009fast}. 
So, the total complexity of our framework for each camera pair is $\mathcal O(n \mbox{log} n)+\mathcal O(m^3)$. It may be noted that $m<<n$ in a practical scenario. In Section \ref{sec:label estimation}, we analyze the real time cost of the proposed method on intra- and inter-camera label estimation.

\section{Experimental Results}
\subsection{Experimental Settings}
\subsubsection{Datasets}
We use two publicly available video re-id datasets for experiments such as MARS dataset and DukeMTMC-VideoReID dataset. {\bf MARS} \cite{zheng2016mars} is captured by 6 cameras and contains 20,715 video tracklets of 1,261 identities. 
{\bf DukeMTMC-VideoReID} (Duke-VideoReID) \cite{wu2018exploit} is from the DukeMTMC dataset \cite{ristani2016performance} which is captured by 8 cameras and contains 4,832 tracklets of 1,404 identities.
It may be noted that the proposed global camera network constraints are based on the triangle relationships in a camera network. That means we need the matching associations between every two of the three cameras at each time. Therefore, the proposed method can be used to the scenarios where there are more than two cameras in a camera network.

\begin{algorithm}[t]
  \caption{Consistent Cross-view Matching}
  \label{alg::ccm}
  \begin{algorithmic}[1]
    \Require Samples in each camera: $\mathcal I_r=\left\{I_r^1,I_r^2,...,I_r^{N_r}\right\}$; pre-trained feature extractor: $f(\cdot)$;
    initialized distance metric model $M_{p,q}^0=I$ for camera pair $(p,q)$; the number of iterations: {\it maxIter}.
    \Ensure
      Distance metric models $M_{p,q}$ for camera pairs.
    \State $f(\cdot): \mathcal I_r \rightarrow \mathcal T_r=\left\{T_r^1,T_r^2,...,T_r^{N_r}\right\}$;    //Feature extraction;
    \State {\bf {Intra-camera Relationships Explorations}}:
     \State $\mathcal T_r \rightarrow \mathcal C_r=\left\{\mathcal C_r^1,\mathcal C_r^2,...,\mathcal C_r^{n_r}\right\}$; // intra-camera clustering with Eq. (1);
     \State {\bf {Inter-camera Relationships Explorations}}:
      \State construct graphs $\mathcal G_{p,q}=(\mathcal C_p,\mathcal C_q,E_{M_{p,q}^0})$ for camera pairs;
      \State compute the assignment matrix $X_{p,q}^0$ with Eq. (5);
      \State compute the consistent matches $\hat X_{p,q}^0$ with Eq. (8-9); // Global camera network constraints
    \For {$t=1$ to $maxIter$}
      \State update the distance metric model $M_{p,q}^t$ with Eq. (11);
      \State update assignment matrix $X_{p,q}^t$ with Eq. (5);
            \If {$G(X_{p,q}^t;M_{p,q}^t) \leq G(X_{p,q}^{t-1};M_{p,q}^t)$}
            \State break;
            \EndIf
      \State update the consistent matches $\hat{X}_{p,q}^t$ with Eq. (8-9);
    \EndFor
  \end{algorithmic}
\end{algorithm}

\subsubsection{Feature extraction}
In the training stage, we first employ a pre-trained feature embedding model to extract features for the training samples and use them as the inputs of our approach. In this paper, both hand-crafted features (LOMO) \cite{liao2015person} and deep convolutional neural network (CNN) features are considered for evaluating the performance of our proposed method. The LOMO feature descriptor is of 26,960 dimensions and we use the principal component analysis (PCA) method \cite{wold1987principal} to reduce the dimension to 600. We adopt the pre-trained unsupervised feature embedding model in \cite{lin2019bottom} which is designed for unsupervised person re-id to extract the deep CNN features and then $\ell_2 $ normalize it for all experiments. For the video-based datasets, we conduct mean-pooling for each tracklet to get more robust video feature representations.

\subsubsection{Evaluation metrics}
We follow the standard training/testing split \cite{zheng2016mars, wu2018exploit} of the two datasets to train and test the proposed model. The Cumulative Matching Characteristic (CMC) curve and the mean average precision (mAP) are utilized to evaluate the performance of each method. 

\subsubsection{Implementation details}
In this paper, we employ Hungarian algorithm \cite{kuhn1955hungarian} to solve the binary  linear  programming  problem and the log-logistic metric learning (MLAPG) \cite{liao2015efficient} for matching persons in re-id. Note that our approach is not specific to any type of matching and metric learning algorithms used for person re-id. During training, we learn a metric model for each pair of cameras using their corresponding matched pairs, and in testing, we match each query-gallery pair using the learned metric models and take the minimum value for each pair. 
All the reported results are based on $\theta=1$  and deep CNN features. Moreover, on the MARS dataset, we evaluate the performance of the proposed method using both LOMO and CNN features. Note that our method does not require any labeled samples during model training and on the DukeMTMC-VideoReID dataset, we conduct cross-view matching directly without intra-camera clustering as the number of samples captured by each camera is small. The specific training process of the proposed consistent cross-view matching method can be found in Algorithm 1.

\begin{table}
\centering
\caption{Rank-1, -5, -10 accuracy (\%) and mAP (\%) performance using different intra-camera clustering methods on the MARS dataset.
}
\small
\setlength{\tabcolsep}{13pt}
\begin{threeparttable}
\begin{tabular}{lllll}
\toprule
MARS & R1 & R5 & R10 & mAP \\
\hline
DBSCAN &60.9 & 72.6 &77.2 &37.4\\
HDBSCAN &62.1 &74.2 &78.5 &38.3\\
OURS + FN-C     & 65.3& 77.8& 81.3& 41.2\\
OURS + Per-C     & 66.0& 77.8& 81.9& 42.3\\
\bottomrule

\end{tabular}
 OURS+ FN-C represents that we use the first neighbor-based strategy for intra-camera samples clustering in our framework; OURS + Per-C denotes that we employ the perfect intra-camera associations in our framework.
\end{threeparttable}
\label{table:clustering}
\end{table}

\subsection{Evaluation of Label Estimation}
\label{sec:label estimation}
In this section, we evaluate the intra- and cross-camera label estimation performance of our proposed method. Specifically, on the MARS dataset, we first measure the intra-camera label estimation performance and observe that 70.9\% samples of the same identities are clustered correctly indicating that our first neighbor-based clustering method is efficient for intra-camera label estimation. In addition, to better evaluate the advantages of the first neighbor-based intra-camera clustering strategy, we compare the recognition performance with other clustering methods, such as DBSCAN \cite{ester1996density} and HDBSCAN \cite{campello2013density,campello2015hierarchical} on the MARS dataset. From Table \ref{table:clustering}, it can be seen that our method outperforms the other clustering methods consistently. Comparing to HDBSCAN, the recognition performance is improved by 3.2\% and 2.9\% for rank-1 accuracy and mAP score, respectively. 
We further evaluate our method with the perfect intra-camera clustering performance (Per-C) which means that samples with the same identity are grouped together manually in each camera. It can be seen from Table \ref{table:clustering} that by using the perfect intra-camera associations, the rank-1 accuracy and mAP score are just increased by 0.7\%  and 1.1\% compared to the first neighbor based clustering strategy (FN-C). We may conclude that the first neighbor based clustering method is effective for intra-camera associations exploration.

We next validate the advantages of the proposed consistent cross-view matching approach for cross-camera label estimation. Specifically, we assume that in each camera, we can group all samples with the same identity together. It may be noted that this assumption just works in this subsection. 

\begin{table}[]
\centering
\caption{Performance of cross-view matching with/without global camera network constraints on two datasets.}
\small
\setlength{\tabcolsep}{6pt}
\begin{threeparttable}
\begin{tabular}{lllll}
\toprule
Dataset& Setting& Pr (\%) & Re (\%) & F1 (\%) \\
\hline
\multirow{2}{*}{MARS} & w/o GNC& 59.5& 82.7& 69.2\\ & w/ GNC& 72.3& 76.8& 74.5\\

\hline
\multirow{2}{*}{Duke-VideoReID}& w/o GNC& 21.7& 81.5& 34.3\\  & w/ GNC& 50.9 &55.0 & 52.9\\

\bottomrule
\end{tabular}
w/o GNC: cross-view matching without global network constraints; w/ GNC: cross-view matching with global network constraints.

\end{threeparttable}
\label{table:cross-view}
\end{table}

On top of the perfect intra-camera clustering results, Table \ref{table:cross-view} reports the performance of cross-camera label estimation with or without global network constraints (GNC). The standard precision (Pr), recall (Re) and F1-score (F1) are utilized to illustrate the performance of the proposed consistent cross-view matching approach across a camera network. We can see that by introducing global network constraints the precision score is improved by a large margin, especially, the improvement of 12.8\% and 29.2\% can be obtained on MARS and DukeMTMC-VideoReID datasets, respectively and on the MARS dataset, 72.3\% matched pairs are the correct matches. Moreover, it can be observed that by introducing the global network constraints into cross-view matches, the recall value drops a lot, but the F1-score is significantly improved. We believe this is due to Hungarian algorithm tries to obtain as many matching associations as possible across camera pairs and hence produces many false positive pairs. 

\begin{figure}
\centerline{\includegraphics[width=1\columnwidth]{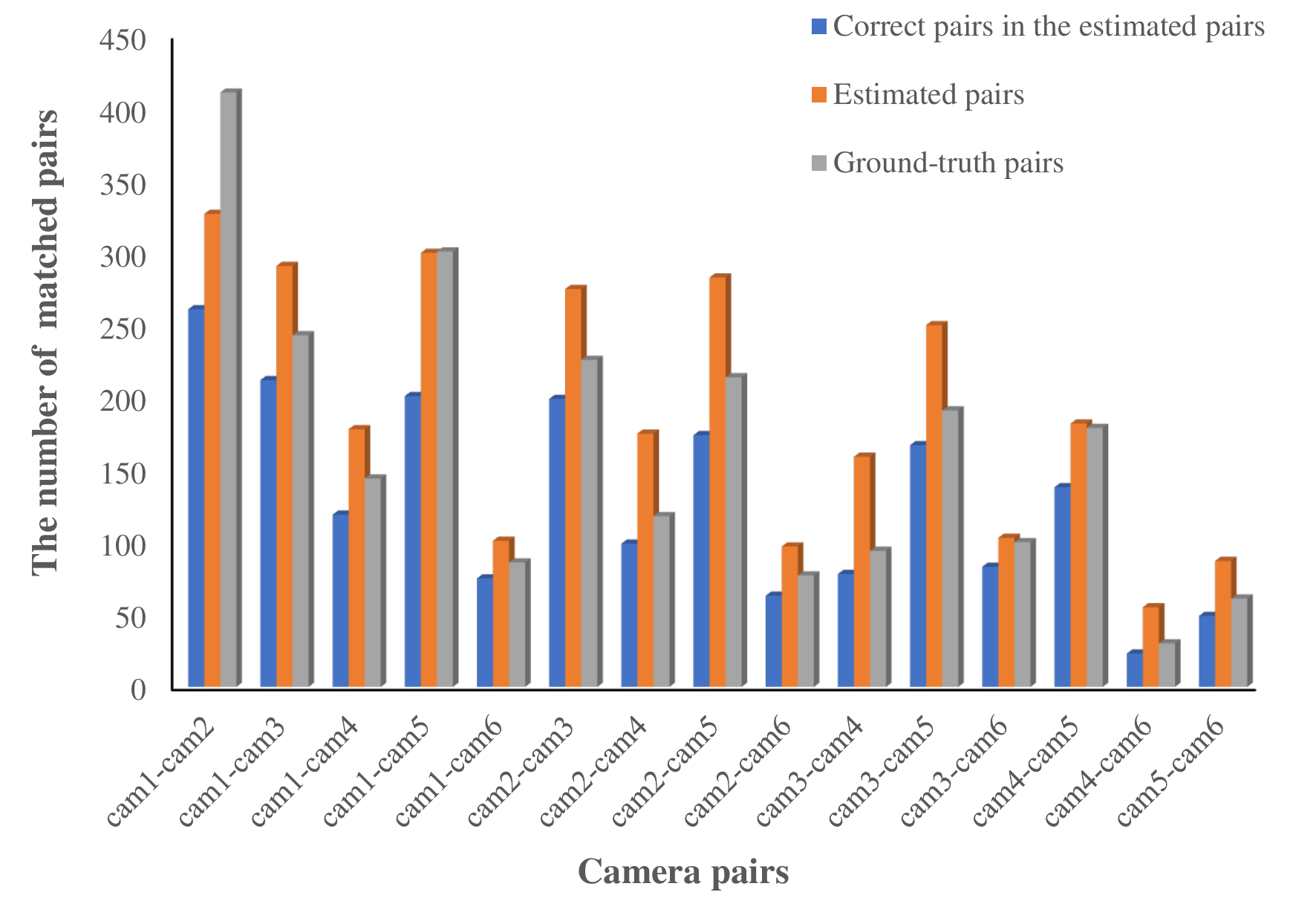}}
\caption{The number of cross-camera matches on the MARS dataset.}
\label{fig:cross-camera}
\end{figure}

We further demonstrate some specific cross-camera label estimation results on the MARS dataset as shown in Figure \ref{fig:cross-camera}. In the figure, it can be seen that the proposed method obtains most of the correct pairs and reject the false positive matches across all the pairs of cameras. For example, camera \#1 captures 520 different persons, camera \#6 captures 104 persons on the MARS dataset, and there are 86 persons in common. That means 452 outliers will affect their matching performance. However, Figure \ref{fig:cross-camera} shows us that 101 matches can be obtained by our method, including 75 correct pairs among all 86 ground-truth matches. This again demonstrates that our method is very effective for cross-camera label estimation in a network of cameras. 

We also measure the real time cost of our method on label estimation. Specifically, on the MARS dataset, for each camera, it takes 0.27 seconds for intra-camera label estimation on average and it takes an average of 45.12 seconds for cross-camera label estimation. It may be noted that all experiments are conducted on an i5-7200U CPU.

\begin{table}[]
\small
\centering
\caption{Rank-1 accuracy (\%) and mAP score (\%) on different $\theta$ Values.}
\begin{tabular}{lllllll}
\toprule
\multicolumn{1}{l}{\multirow{2}{*}{Dataset}} & 
\multicolumn{2}{c}{$\theta = 0$} & \multicolumn{2}{c}{$\theta=1$} & \multicolumn{2}{c}{$\theta = 2$} \\ 
\cline{2-7} 
\multicolumn{1}{c}{} & R1 & mAP& \multicolumn{1}{l}{R1} & mAP & R1 & mAP   \\ 
\hline
MARS & \multicolumn{1}{l}{63.7} & 35.7 & 65.3 & 41.2 & 61.1 & 38.5  \\

Duke-VideoReID & \multicolumn{1}{l}{73.1} & 63.9 & 76.5 & 68.7  & 75.4 & 67.0  \\ 
\bottomrule
\end{tabular}
\label{table:theta}
\end{table}

\subsection{Evaluation of Different Reliability Values $\theta$}
{The quality and quantity of the estimated matching pairs are very important for learning an efficient pair-wise metric models in unsupervised person re-id. Both of them are related to the reliability value $(\theta)$ of the matches. Thus, we compare the recognition performance under different $\theta$ values to select the optimal one. Table~\ref{table:theta} shows the Rank-1 accuracy and mAP score under 3 different reliability values. 
We can see that the recognition performance fluctuates a little with the increase in $\theta$ values because of the trade-off between the quantity and reliability of the matched pairs. A small reliability value means that we can collect most of cross-view matches, but the learned metric models using these pairs may not perform well because it also introduces massive false positive pairs. As the $\theta$ value increases, the number of matched pairs will decline, but the reliability of the matches will increase. We observe that when $\theta=1$, we obtain the best recognition performance with 65.3\% 
and 41.2\% for rank-1 accuracy and mAP score respectively on the MARS dataset, similarly 76.5\% and 68.7\% on the DukeMTMC-VideoReID dataset.
} 

\begin{table}
\centering
\caption{Rank-1, -5, -10 accuracy (\%) and mAP (\%) performance using cross-view distance metric models.
}
\small
\setlength{\tabcolsep}{9pt}
\begin{threeparttable}
\begin{tabular}{lllll}
\toprule
MARS & R1& R5& R10& mAP\\
\hline
OURS w/o cross-view& 65.2& 77.1& 81.1& 40.4\\
OURS w cross-view& 65.3& 77.8& 81.3& 41.2\\
\hline

\hline
DukeMTMC-VideoReID& R1& R5& R10& mAP\\
\hline
OURS w/o cross-view& 73.6& 87.1& 90.2& 65.5\\
OURS w cross-view& 76.5& 89.6& 91.9& 68.7\\
\bottomrule

\end{tabular}
 w cross-view: training a metric model for each camera pair. w/o cross-view: training a global metric model for the entire dataset.
\end{threeparttable}
\label{table:metric}
\end{table}

\subsection{Evaluation of Cross-view Metric Model}
In the proposed consistent cross-view matching framework, we train a separate metric model for each pair of cameras. In this section, we show the effectiveness of the cross-view metric models over a global metric model for the entire dataset.

From Table \ref{table:metric}, it can be seen that the cross-view metric models perform better than that training a global metric model for the entire dataset. Specifically, on the MARS dataset, the rank-1 accuracy and mAP score are improved by 0.1\% and 0.8\%, respectively. On the DukeMTMC-VideoReID dataset, the rank-1 accuracy is increased from 73.6\% to 76.5\% (1.8\% difference), and from 65.5\% to 68.7\% (1.5\% difference) for mAP score.

\begin{figure}
\centerline{\includegraphics[width=1\columnwidth]{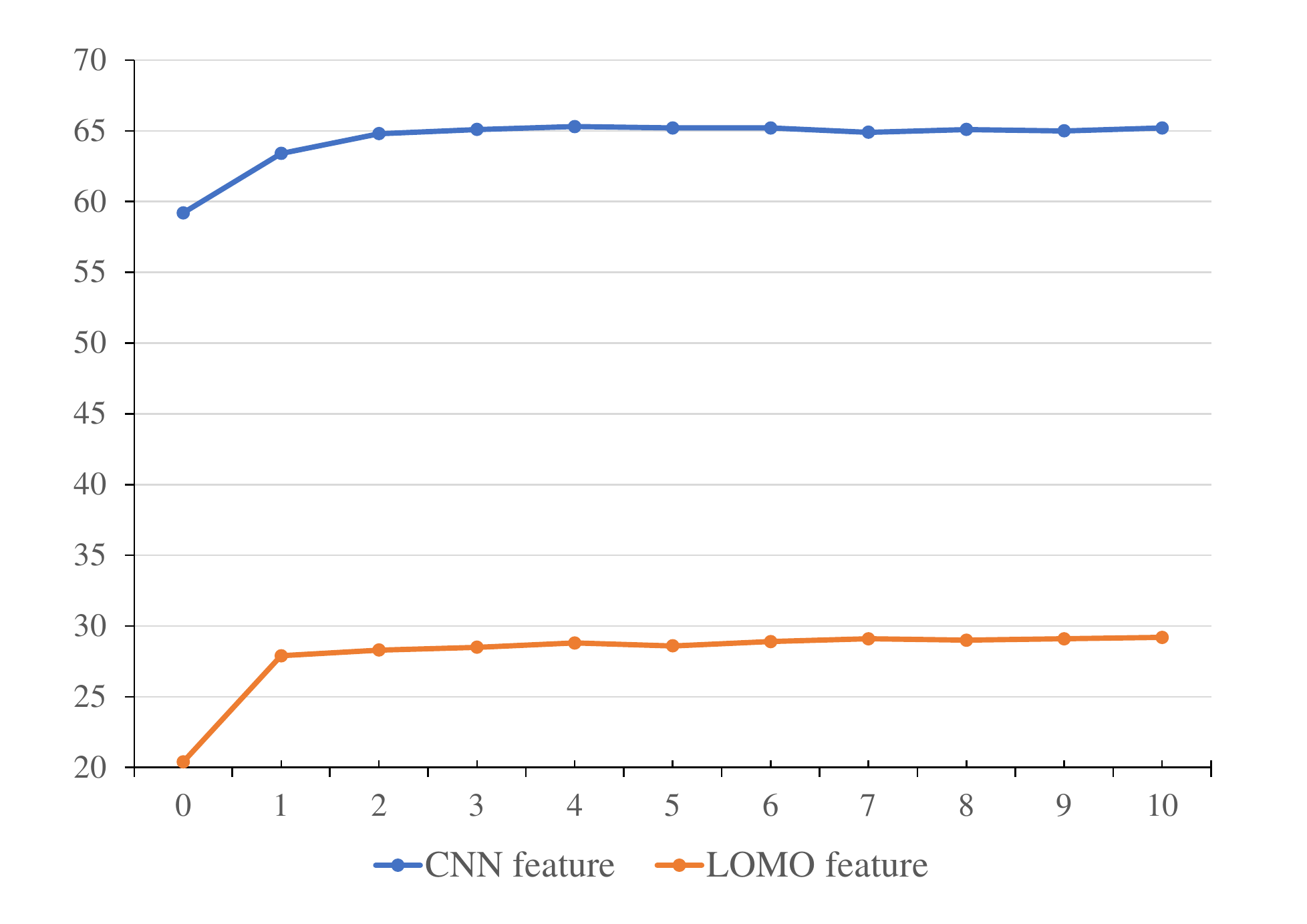}}
\setlength{\abovecaptionskip}{0.cm}
\caption{Rank-1 accuracy of the proposed method on the MARS dataset with the LOMO feature and CNN feature at each iteration.}
\label{fig:iteration}
\end{figure}

\subsection{Evaluation of Iterative Updating}
Figure \ref{fig:iteration} shows the rank-1 accuracy of the proposed method using LOMO feature and CNN feature on the MARS dataset at each iteration. 
We iterate each experiment 10 times. As may be observed from the plot, with the increase in the number of iterations, the rank-1 accuracy is improved from 59.2\% to 65.3\% for CNN feature and 20.4\% to 29.2\% for LOMO feature, and after that, the plots are almost stable. By introducing the conditions of iteration end, our experiments will end at the 4th iteration and the 7th iteration for CNN feature and LOMO feature, respectively.

\begin{table}
\centering
\caption{Rank-1, -5, -10 accuracy (\%) and mAP (\%) with some
unsupervised and semi-supervised approaches on the MARS dataset.}
\small
\begin{threeparttable}
\begin{tabular}{llllll}

\toprule
Methods & Labels & R1& R5& R10& mAP\\
\hline
GRDL \cite{kodirov2016person}& None & 19.3& 33.2& 41.6& 9.6\\
UnKISS \cite{khan2016unsupervised}& None& 22.3& 37.4& 47.2& 10.6\\
DGM+LOMO \cite{ye2019dynamic}& None& 24.7& 39.4& 47.0& 11.7\\
\bf{OURS+LOMO}& None& \bf{29.2}& \bf{44.3}& \bf{50.5}& \bf{12.2}\\\hline
OIM \cite{xiao2017joint}& None& 33.7& 48.1& 54.8& 13.5\\
UTM \cite{riachy2019video} & None& 39.7& 53.2& -& 20.1\\ 
DGM+IDE \cite{ye2019dynamic}& None& 48.1& 64.7& 71.1& 29.2\\
DAL \cite{chen2018bmvc}& None& 49.3& 65.9& 72.2& 23.0\\
BUC \cite{lin2019bottom}& None& 61.1& 75.1& 80.0& 38.0\\
TAULD \cite{li2018unsupervised}&None &43.8 &59.9 &72.8 &29.1\\
UTAL \cite{li2019Unsupervised}&None & 49.9 &66.4&77.8 &35.2\\
TSSL \cite{wu2020tracklet} &None & 56.3&-&-& 30.5\\
CCE \cite{lin2020unsupervised} &None &62.8 &77.2 &80.1 &\bf{43.6}\\
\bf OURS& None& \bf{65.3}& \bf{77.8}& \bf 81.3& 41.2\\
\hline
UGA \cite{wu2019unsupervised}&Intra-camera & 59.9 &- &- &40.5\\
Prog. Learning \cite{wu2019progressive}&One-shot &62.8 &75.2 &80.4 & 42.6\\
Stepwise \cite{liu2017stepwise}& One-shot& 41.2& 55.5&-& 19.6\\
RACE \cite{ye2018race}& One-shot& 43.2& 57.1& 62.1& 24.5\\
EUG \cite{wu2018exploit}& One-shot&62.6 &74.9 &\bf 82.5 &42.4\\
TCPL \cite{raychaudhuri2020exploiting} & One-shot & 65.2 &77.5 &- &\bf 43.6\\
\textbf{OURS} & None& \bf 65.3& \bf 77.8&  81.3& 41.2\\
\bottomrule
\end{tabular}
None denotes fully unsupervised methods; One-shot assumes a singular labeled tracklet for each identity along with a large pool of unlabeled samples; Intra-camera setting works with labels which are provided only for samples within an individual camera view. 
\end{threeparttable}
\label{table:mars}
\end{table}

\subsection{Comparison to the SOTA Methods}
{\bf{MARS Dataset.}}
We compare our approach with several state-of-the-art person re-id methods that fall into two main categories: \emph {unsupervised methods} such as GRDL \cite{kodirov2016person}, UnKISS \cite{khan2016unsupervised}, DGM+ \cite{ye2019dynamic} using LOMO feature, DGM+ \cite{ye2019dynamic} using deep IDE features, OIM \cite{xiao2017joint}, DAL \cite{chen2018bmvc}, BUC \cite{lin2019bottom},  UTM \cite{riachy2019video}, TAULD \cite{li2018unsupervised}, UTAL \cite{li2019Unsupervised}, TSSL \cite{wu2020tracklet}, CCE \cite{lin2020unsupervised} and \emph{semi-supervised methods} such as UGA (intra-camera supervision) \cite{wu2019unsupervised}, Progressive Learning (one-shot setting) \cite{wu2019progressive}, Stepwise (one-shot setting) \cite{liu2017stepwise}, RACE (one-shot setting) \cite{ye2018race} and EUG (one-shot setting) \cite{wu2018exploit}. 
As seen from Table \ref{table:mars}, while comparing with unsupervised alternatives, we evaluate our method in two different settings: (1) methods based on hand-crafted features: the proposed method significantly outperforms all the compared methods; comparing to DGM+, we achieve 4.5\% and 0.5\% improvement using the same LOMO feature in rank-1 accuracy and mAP score, respectively; (2) methods based on deep learning: our method also obtains the best recognition performance 65.3{\%} for rank-1 and 41.2{\%} for mAP while comparing to fully unsupervised deep learning based alternatives, especially, comparing to BUC, the rank-1 accuracy and mAP score are improved by 4.2\% and 3.2\%, respectively.Compared with the recent CCE \cite{lin2020unsupervised}, our method improves the rank-1 accuracy from 62.8\% to 65.3\%.
As expected, the proposed method performs better while using the deep CNN features compared to the handcrafted LOMO features. 
Moreover, from Table \ref{table:mars}, we observe that the proposed method is also very competitive while comparing with the semi-supervised methods without requiring any person identity information. Comparing to EUG (one-shot setting), our method achieves 2.7\% improvement in rank-1 accuracy.
It may be noted that any unsupervised feature embedding learning-based person re-id models \cite{lin2019bottom,lin2020unsupervised} can be used as our feature extractors, and the better the feature extractors are, the better our method performs. 

\begin{table}
\centering
\caption{Rank-1, -5, -10 accuracy (\%) and mAP (\%) with some
unsupervised and one-shot supervised approaches on
the DukeMTMC-VideoReID dataset.}

\small
\begin{tabular}{llllll}
\toprule
Methods& Labels&R1& R5& R10& mAP\\
\hline
OIM \cite{xiao2017joint}& None& 51.1& 70.5& 76.2& 43.8\\
DGM+IDE \cite{ye2019dynamic}& None &42.3& 57.9& 69.3& 33.6\\
TAUDL \cite{li2018unsupervised}&None&26.1&42.0&57.2&20.8\\
UTAL \cite{li2019Unsupervised}&None& 48.3&62.8&76.5&36.6\\
TSSL\cite{wu2020tracklet}&None&73.9&-&-&64.6\\
BUC \cite{lin2019bottom}& None& 69.2& 81.1& 85.8& 61.9\\
CCE \cite{lin2020unsupervised} &None &76.4 &88.7 &91.0 &\bf{69.3}\\
\textbf{OURS} & None& {\bf 76.5}&{\bf 89.6}&{\bf 91.9}&68.7\\
\hline
Stepwise \cite{liu2017stepwise}& One-shot& 56.2& 70.3& 79.2& 46.7\\
EUG \cite{wu2018exploit}&One-shot& 72.7& 84.1& -& 63.2\\
Prog. Learning \cite{wu2019progressive}&One-shot &72.9&84.3&88.3&63.3\\
\textbf{OURS} & None& {\bf 76.5}&{\bf 89.6}&{\bf 91.9}&{\bf 68.7}\\
\bottomrule
\end{tabular}

\label{table:dukeV}
\end{table}

\begin{table}
\centering
\caption{Ablation studies of the proposed method on the MARS and DukeMTMC-VideoReID datasets. Rank-1,-5,-10
accuracies(\%) and mAP (\%) are reported
}
\small
\setlength{\tabcolsep}{9pt}
\begin{threeparttable}
\begin{tabular}{lllll}
\toprule
MARS & R1& R5& R10& mAP\\
\hline
Baseline& 61.1& 75.1& 80.0& 38.0\\
OURS w/ CM& 64.6& 76.7& 80.7& 39.8\\
OURS w/ CM+GNC& 65.3& 77.8& 81.3& 41.2\\
\hline
\hline
DukeMTMC-VideoReID & R1& R5& R10& mAP\\
\hline
Baseline& 69.2& 81.1& 85.8& 61.9\\
OURS w/ CM& 72.2& 86.2& 89.3& 64.3\\
OURS w/ CM+GNC& 76.5& 89.6& 91.9& 68.7\\
\bottomrule
\end{tabular}
Baseline: Recognition performance is measured by directly using the Euclidean distance ($\ell_2 $ distance). w/ CM: Cross-view matching by introducing the graph matching into baseline. GNC: global network constraints.
\end{threeparttable}
\label{table:ablation}
\end{table}

{\bf {DukeMTMC-VideoReID Dataset.}}
We also evaluate our method on a larger video person re-id dataset - DukeMTMC-VideoReID dataset which is captured with 8 different cameras by comparing with several state-of-the-art methods such as OIM \cite{xiao2017joint}, DGM+ \cite{ye2019dynamic}, Stepwise \cite{liu2017stepwise}, EUG \cite{wu2018exploit}, Progressive Learning \cite{wu2019progressive}, BUC \cite{lin2019bottom}, TAULD \cite{li2018unsupervised}, UTAL \cite{li2019Unsupervised}, TSSL \cite{wu2020tracklet} and CCE \cite{lin2020unsupervised}. Results in Table \ref{table:dukeV} shows the superiority of the proposed unsupervised framework over all the compared methods (unsupervised or one-shot supervised methods). We achieve the best recognition performance with rank-1 accuracy of 76.5\% and mAP score of 68.7\%, respectively. Comparing to BUC (unsupervised), the proposed method achieves 7.2\% and 6.8\% improvement for rank-1 accuracy and mAP score, respectively. Comparing EUG (one-shot setting), the recognition performance is improved from 72.7\% to 76.5\% for rank-1 accuracy and 63.2\% to 68.7\% for mAP score, respectively.

\subsection{Ablation Studies}
To better evaluate the effectiveness of our proposed method, we conduct ablation studies on the DukeMTMC-VideoReID dataset and MARS dataset. 
As shown in Table \ref{table:ablation}, Baseline denotes that we measure the recognition performance using the Euclidean distance ($\ell_2 $ distance) directly on the extracted features.
We first show the effect caused by the cross-view matching (CM) via introducing the graph matching into the Baseline. 
On the DukeMTMC-VideoReID dataset, it can be seen that ``Ours w/ CM" improves the recognition performance from 69.2\% to 72.2\% for rank-1 accuracy and 61.9\% to 64.3\% for mAP score, similarity, 3.5\% and 1.8\% improvement on the MARS dataset. This demonstrates that the cross-view matching is an effective way to improve the performance by exploiting the similarity relationships across camera pairs. Furthermore, we validate the effect caused by the global network constraints (GNC). As shown in Table \ref{table:ablation}, by introducing the GNC, the recognition performance is improved consistently.
``Ours w/ CM+GNC" achieves the best recognition performance with rank-1 accuracy of 76.5\% and mAP score of 68.7\% on the DukeMTMC-VideoReID dataset, and 65.3\% and 41.2\% on the MARS dataset.

\begin{figure}
\centering
\centerline{\includegraphics[width=\columnwidth]{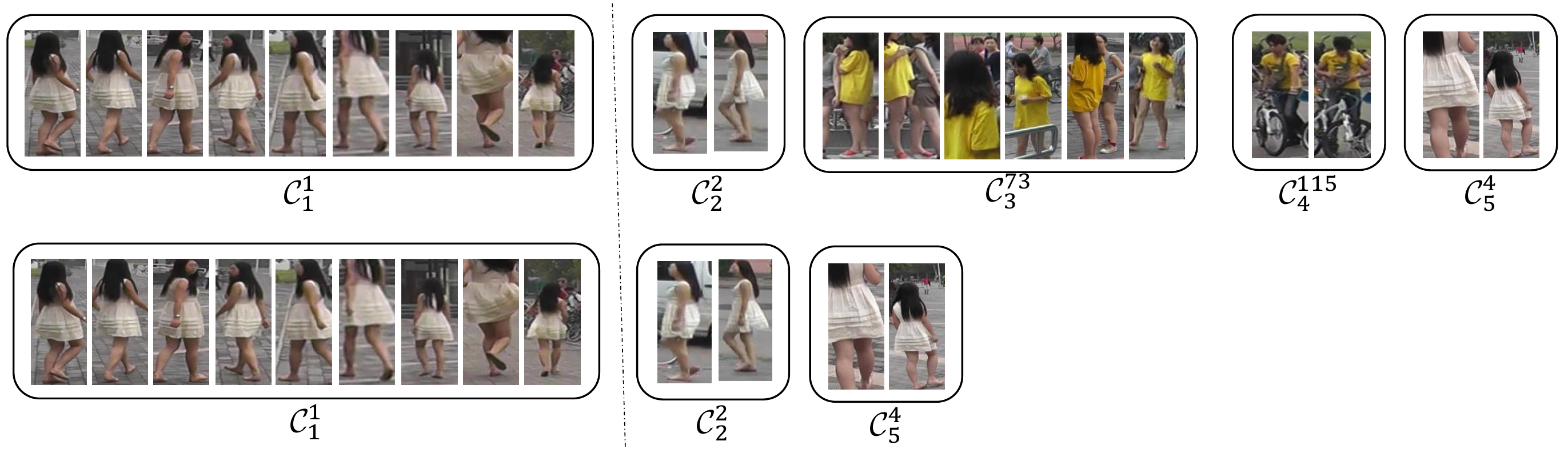}}
\caption{Visualization of intra-camera and inter-camera label estimation with/without global network constraints on the MARS dataset. Samples in each box denote the intra-camera clustering results and  $\mathcal C_r^i$ represents the $i$th cluster in camera $r$. The first and second rows show the cross-camera matching performance without and with global network constraints, respectively. Note that each image in this figure denotes one person tracklet and we illustrate the matching performance with one person/cluster.}
\label{fig:visualization}
\end{figure}

The visualization of intra-camera and inter-camera label estimation results with/without global network constraints on MARS is shown in
Figure~\ref{fig:visualization}. It can be seen that the first neighbor-based clustering strategy is effective to group the similar samples in each camera. When introducing global network constraints, the outliers can be removed significantly. Note that we set $\theta=1$ and $\mbox{iteration}=0$ to show the results of cross-camera label estimation.

\section{Conclusions and Future Works}
In this paper, we propose a fully unsupervised consistent cross-view matching framework for video-based person re-identification. We first propose to use first neighbor of each sample to explore the correlations in each camera and then global camera network constraints are introduced into cross-view matches to reduce the wrong matches significantly. We then present a definition of cross-view matching reliability, which can be used to balance the performance between the quality and quantity of the estimated pairs. In addition, our consistent cross-view matching method is embedded into an iterative framework which iterates between the cross-camera label estimation and metric models learning. Rigorous experiments on two standard video person re-id datasets show the advantages of our approach over the state-of-the-art methods.

The proposed method learns metric models for camera pairs progressively, which relies on the pre-trained feature embedding models. In the future, we will try to design an end-to-end model that jointly optimizes the feature extractor and distance metric models by mining the consistent cross-camera pairs for person re-id task.

\section*{Acknowledgment}
This work was supported in part by the National Natural Science Foundation of China under Grant No. 61771189 and the Hunan Provincial Natural Science Foundation of China under Grant No.2018JJ2060, in part by ONR grant N00014-19-1-2264 and NSF grant 1544969, and in part by China Scholarship Council.

\bibliographystyle{IEEEtran}  
\bibliography{reference.bib}  

\begin{thebibliography}{10}
\providecommand{\url}[1]{#1}
\csname url@samestyle\endcsname
\providecommand{\newblock}{\relax}
\providecommand{\bibinfo}[2]{#2}
\providecommand{\BIBentrySTDinterwordspacing}{\spaceskip=0pt\relax}
\providecommand{\BIBentryALTinterwordstretchfactor}{4}
\providecommand{\BIBentryALTinterwordspacing}{\spaceskip=\fontdimen2\font plus
\BIBentryALTinterwordstretchfactor\fontdimen3\font minus
  \fontdimen4\font\relax}
\providecommand{\BIBforeignlanguage}[2]{{%
\expandafter\ifx\csname l@#1\endcsname\relax
\typeout{** WARNING: IEEEtran.bst: No hyphenation pattern has been}%
\typeout{** loaded for the language `#1'. Using the pattern for}%
\typeout{** the default language instead.}%
\else
\language=\csname l@#1\endcsname
\fi
#2}}
\providecommand{\BIBdecl}{\relax}
\BIBdecl

\bibitem{roy2012camera}
A.~K. Roy-Chowdhury and B.~Song, ``Camera networks: The acquisition and
  analysis of videos over wide areas,'' \emph{Synthesis Lectures on Computer
  Vision}, vol.~3, no.~1, pp. 1--133, 2012.

\bibitem{arxiv20reidsurvey}
M.~Ye, J.~Shen, G.~Lin, T.~Xiang, L.~Shao, and S.~C.~H. Hoi, ``Deep learning
  for person re-identification: A survey and outlook,'' \emph{arXiv preprint
  arXiv:2001.04193}, 2020.

\bibitem{chen2019spatial}
G.~Chen, J.~Lu, M.~Yang, and J.~Zhou, ``Spatial-temporal attention-aware
  learning for video-based person re-identification,'' \emph{IEEE Transactions
  on Image Processing}, vol.~28, no.~9, pp. 4192--4205, 2019.

\bibitem{rao2019learning}
Y.~Rao, J.~Lu, and J.~Zhou, ``Learning discriminative aggregation network for
  video-based face recognition and person re-identification,''
  \emph{International Journal of Computer Vision}, vol. 127, no. 6-7, pp.
  701--718, 2019.

\bibitem{ye2018race}
M.~Ye, X.~Lan, and P.~C. Yuen, ``Robust anchor embedding for unsupervised video
  person re-identification in the wild,'' in \emph{Proceedings of European
  Conference on Computer Vision}, 2018, pp. 170--186.

\bibitem{ye2019dynamic}
M.~{Ye}, J.~{Li}, A.~J. {Ma}, L.~{Zheng}, and P.~C. {Yuen}, ``Dynamic graph
  co-matching for unsupervised video-based person re-identification,''
  \emph{IEEE Transactions on Image Processing}, vol.~28, no.~6, pp. 2976--2990,
  2019.

\bibitem{li2019global}
J.~Li, J.~Wang, Q.~Tian, W.~Gao, and S.~Zhang, ``Global-local temporal
  representations for video person re-identification,'' in \emph{Proceedings of
  the IEEE International Conference on Computer Vision}, 2019, pp. 3958--3967.

\bibitem{ouyang2018video}
D.~Ouyang, J.~Shao, Y.~Zhang, Y.~Yang, and H.~T. Shen, ``Video-based person
  re-identification via self-paced learning and deep reinforcement learning
  framework,'' in \emph{Proceedings of ACM International Conference on
  Multimedia}, 2018, pp. 1562--1570.

\bibitem{ye2020purifynet}
M.~Ye and P.~C. Yuen, ``Purifynet: A robust person re-identification model with
  noisy labels,'' \emph{IEEE Transactions on Information Forensics and
  Security}, vol.~15, pp. 2655--2666, 2020.

\bibitem{chen2018bmvc}
Y.~Chen, X.~Zhu, and S.~Gong, ``Deep association learning for unsupervised
  video person re-identification,'' in \emph{Proceedings of the British Machine
  Vision Conference}, 2018.

\bibitem{li2020hierarchical}
P.~Li, P.~Panb, P.~Liuc, M.~Xu, and Y.~Yang, ``Hierarchical temporal modeling
  with mutual distance matching for video based person re-identification,''
  \emph{IEEE Transactions on Circuits and Systems for Video Technology}, 2020.

\bibitem{ye2017dynamic}
M.~Ye, A.~J. Ma, L.~Zheng, J.~Li, and P.~C. Yuen, ``Dynamic label graph
  matching for unsupervised video re-identification,'' in \emph{Proceedings of
  the IEEE International Conference on Computer Vision}, 2017, pp. 5142--5150.

\bibitem{riachy2019video}
C.~Riachy, F.~Khelifi, and A.~Bouridane, ``Video-based person re-identification
  using unsupervised tracklet matching,'' \emph{IEEE Access}, vol.~7, pp.
  20\,596--20\,606, 2019.

\bibitem{wu2020tracklet}
G.~Wu, X.~Zhu, and S.~Gong, ``Tracklet self-supervised learning for
  unsupervised person re-identification.'' in \emph{Proceedings of AAAI
  Conference on Artificial Intelligence}, 2020, pp. 12\,362--12\,369.

\bibitem{tip20mace}
M.~Ye, X.~Lan, Q.~Leng, and J.~Shen, ``Cross-modality person re-identification
  via modality-aware collaborative ensemble learning,'' \emph{IEEE Transactions
  on Image Processing (TIP)}, 2020.

\bibitem{wang2020learning}
X.~Wang, S.~Paul, D.~S. Raychaudhuri, M.~Liu, Y.~Wang, A.~K. Roy-Chowdhury
  \emph{et~al.}, ``Learning person re-identification models from videos with
  weak supervision,'' \emph{arXiv preprint arXiv:2007.10631}, 2020.

\bibitem{pami20embedding}
M.~Ye, J.~Shen, X.~Zhang, P.~C. Yuen, and S.-F. Chang, ``Augmentation invariant
  and instance spreading feature for softmax embedding,'' \emph{IEEE TPAMI},
  2020.

\bibitem{lin2019bottom}
Y.~Lin, X.~Dong, L.~Zheng, Y.~Yan, and Y.~Yang, ``A bottom-up clustering
  approach to unsupervised person re-identification,'' in \emph{Proceedings of
  the AAAI Conference on Artificial Intelligence}, 2019, pp. 8738--8745.

\bibitem{fan2018unsupervised}
H.~Fan, L.~Zheng, C.~Yan, and Y.~Yang, ``Unsupervised person re-identification:
  Clustering and fine-tuning,'' \emph{{ACM} Transactions on Multimedia
  Computing, Communications, and Applications}, vol.~14, no.~4, pp.
  83:1--83:18, 2018.

\bibitem{yu2019unsupervised}
H.-X. Yu, W.-S. Zheng, A.~Wu, X.~Guo, S.~Gong, and J.-H. Lai, ``Unsupervised
  person re-identification by soft multilabel learning,'' in \emph{Proceedings
  of the IEEE Conference on Computer Vision and Pattern Recognition}, 2019, pp.
  2148--2157.

\bibitem{fu2019self}
Y.~Fu, Y.~Wei, G.~Wang, Y.~Zhou, H.~Shi, and T.~S. Huang, ``Self-similarity
  grouping: A simple unsupervised cross domain adaptation approach for person
  re-identification,'' in \emph{Proceedings of the IEEE International
  Conference on Computer Vision}, 2019, pp. 6112--6121.

\bibitem{zhang2019self}
X.~Zhang, J.~Cao, C.~Shen, and M.~You, ``Self-training with progressive
  augmentation for unsupervised cross-domain person re-identification,'' in
  \emph{Proceedings of the IEEE International Conference on Computer Vision},
  2019, pp. 8222--8231.

\bibitem{zhang2016prism}
Z.~Zhang and V.~Saligrama, ``Prism: Person reidentification via structured
  matching,'' \emph{IEEE Transactions on Circuits and Systems for Video
  Technology}, vol.~27, no.~3, pp. 499--512, 2016.

\bibitem{chu2014fully}
C.-T. Chu and J.-N. Hwang, ``Fully unsupervised learning of camera link models
  for tracking humans across nonoverlapping cameras,'' \emph{IEEE Transactions
  on Circuits and Systems for Video Technology}, vol.~24, no.~6, pp. 979--994,
  2014.

\bibitem{lin2020unsupervised}
Y.~Lin, L.~Xie, Y.~Wu, C.~Yan, and Q.~Tian, ``Unsupervised person
  re-identification via softened similarity learning,'' in \emph{Proceedings of
  the IEEE/CVF Conference on Computer Vision and Pattern Recognition}, 2020,
  pp. 3390--3399.

\bibitem{lin2017consistent}
J.~Lin, L.~Ren, J.~Lu, J.~Feng, and J.~Zhou, ``Consistent-aware deep learning
  for person re-identification in a camera network,'' in \emph{Proceedings of
  the IEEE Conference on Computer Vision and Pattern Recognition}, 2017, pp.
  5771--5780.

\bibitem{das2014consistent}
A.~Das, A.~Chakraborty, and A.~K. Roy-Chowdhury, ``Consistent re-identification
  in a camera network,'' in \emph{Proceedings of European Conference on
  Computer Vision}, 2014, pp. 330--345.

\bibitem{chakraborty2015network}
A.~Chakraborty, A.~Das, and A.~K. Roy-Chowdhury, ``Network consistent data
  association,'' \emph{IEEE Transactions on Pattern Analysis and Machine
  Intelligence}, vol.~38, no.~9, pp. 1859--1871, 2015.

\bibitem{sarfraz2019efficient}
S.~Sarfraz, V.~Sharma, and R.~Stiefelhagen, ``Efficient parameter-free
  clustering using first neighbor relations,'' in \emph{Proceedings of the IEEE
  Conference on Computer Vision and Pattern Recognition}, 2019, pp. 8934--8943.

\bibitem{liny2020unsupervised}
Y.~Lin, Y.~Wu, C.~Yan, M.~Xu, and Y.~Yang, ``Unsupervised person
  re-identification via cross-camera similarity exploration,'' \emph{IEEE
  Transactions on Image Processing}, vol.~29, pp. 5481--5490, 2020.

\bibitem{xiao2017joint}
T.~Xiao, S.~Li, B.~Wang, L.~Lin, and X.~Wang, ``Joint detection and
  identification feature learning for person search,'' in \emph{Proceedings of
  the IEEE Conference on Computer Vision and Pattern Recognition}, 2017, pp.
  3415--3424.

\bibitem{liu2017stepwise}
Z.~Liu, D.~Wang, and H.~Lu, ``Stepwise metric promotion for unsupervised video
  person re-identification,'' in \emph{Proceedings of the IEEE International
  Conference on Computer Vision}, 2017, pp. 2429--2438.

\bibitem{li2018unsupervised}
M.~Li, X.~Zhu, and S.~Gong, ``Unsupervised person re-identification by deep
  learning tracklet association,'' in \emph{Proceedings of the European
  Conference on Computer Vision}, 2018, pp. 737--753.

\bibitem{zhong2018camera}
Z.~Zhong, L.~Zheng, Z.~Zheng, S.~Li, and Y.~Yang, ``Camera style adaptation for
  person re-identification,'' in \emph{Proceedings of the IEEE Conference on
  Computer Vision and Pattern Recognition}, 2018, pp. 5157--5166.

\bibitem{deng2018image}
W.~Deng, L.~Zheng, Q.~Ye, G.~Kang, Y.~Yang, and J.~Jiao, ``Image-image domain
  adaptation with preserved self-similarity and domain-dissimilarity for person
  re-identification,'' in \emph{Proceedings of the IEEE Conference on Computer
  Vision and Pattern Recognition}, 2018, pp. 994--1003.

\bibitem{berg2005shape}
A.~C. Berg, T.~L. Berg, and J.~Malik, ``Shape matching and object recognition
  using low distortion correspondences,'' in \emph{Proceedings of IEEE
  Conference on Computer Vision and Pattern Recognition}.\hskip 1em plus 0.5em
  minus 0.4em\relax IEEE, 2005, pp. 26--33.

\bibitem{yan2015multi}
J.~Yan, M.~Cho, H.~Zha, X.~Yang, and S.~M. Chu, ``Multi-graph matching via
  affinity optimization with graduated consistency regularization,'' \emph{IEEE
  Transactions on Pattern Analysis and Machine Intelligence}, vol.~38, no.~6,
  pp. 1228--1242, 2015.

\bibitem{zhang2016pairwise}
Z.~Zhang, Q.~Shi, J.~McAuley, W.~Wei, Y.~Zhang, and A.~Van Den~Hengel,
  ``Pairwise matching through max-weight bipartite belief propagation,'' in
  \emph{Proceedings of the IEEE Conference on Computer Vision and Pattern
  Recognition}, 2016, pp. 1202--1210.

\bibitem{wu2019unsupervised}
J.~Wu, Y.~Yang, H.~Liu, S.~Liao, Z.~Lei, and S.~Z. Li, ``Unsupervised graph
  association for person re-identification,'' in \emph{Proceedings of the IEEE
  International Conference on Computer Vision}, 2019, pp. 8321--8330.

\bibitem{roy2018exploiting}
S.~Roy, S.~Paul, N.~E. Young, and A.~K. Roy-Chowdhury, ``Exploiting
  transitivity for learning person re-identification models on a budget,'' in
  \emph{Proceedings of the IEEE Conference on Computer Vision and Pattern
  Recognition}, 2018, pp. 7064--7072.

\bibitem{kuhn1955hungarian}
H.~W. Kuhn, ``The hungarian method for the assignment problem,'' \emph{Naval
  Research Logistics Quarterly}, vol.~2, no. 1-2, pp. 83--97, 1955.

\bibitem{liao2015efficient}
S.~Liao and S.~Z. Li, ``Efficient psd constrained asymmetric metric learning
  for person re-identification,'' in \emph{Proceedings of the IEEE
  International Conference on Computer Vision}, 2015, pp. 3685--3693.

\bibitem{tian2012convergence}
Y.~Tian, J.~Yan, H.~Zhang, Y.~Zhang, X.~Yang, and H.~Zha, ``On the convergence
  of graph matching: Graduated assignment revisited,'' in \emph{Proceedings of
  European Conference on Computer Vision}, 2012, pp. 821--835.

\bibitem{beck2009fast}
A.~Beck and M.~Teboulle, ``A fast iterative shrinkage-thresholding algorithm
  for linear inverse problems,'' \emph{SIAM Journal on Imaging Sciences},
  vol.~2, no.~1, pp. 183--202, 2009.

\bibitem{zheng2016mars}
L.~Zheng, Z.~Bie, Y.~Sun, J.~Wang, C.~Su, S.~Wang, and Q.~Tian, ``Mars: A video
  benchmark for large-scale person re-identification,'' in \emph{Proceedings of
  European Conference on Computer Vision}, 2016, pp. 868--884.

\bibitem{wu2018exploit}
Y.~Wu, Y.~Lin, X.~Dong, Y.~Yan, W.~Ouyang, and Y.~Yang, ``Exploit the unknown
  gradually: One-shot video-based person re-identification by stepwise
  learning,'' in \emph{Proceedings of the IEEE Conference on Computer Vision
  and Pattern Recognition}, 2018, pp. 5177--5186.

\bibitem{ristani2016performance}
E.~Ristani, F.~Solera, R.~Zou, R.~Cucchiara, and C.~Tomasi, ``Performance
  measures and a data set for multi-target, multi-camera tracking,'' in
  \emph{Proceedings of European Conference on Computer Vision}, 2016, pp.
  17--35.

\bibitem{liao2015person}
S.~Liao, Y.~Hu, X.~Zhu, and S.~Z. Li, ``Person re-identification by local
  maximal occurrence representation and metric learning,'' in \emph{Proceedings
  of the IEEE Conference on Computer Vision and Pattern Recognition}, 2015, pp.
  2197--2206.

\bibitem{wold1987principal}
S.~Wold, K.~Esbensen, and P.~Geladi, ``Principal component analysis,''
  \emph{Chemometrics and Intelligent Laboratory Systems}, vol.~2, no. 1-3, pp.
  37--52, 1987.

\bibitem{ester1996density}
M.~Ester, H.-P. Kriegel, J.~Sander, X.~Xu \emph{et~al.}, ``A density-based
  algorithm for discovering clusters in large spatial databases with noise.''
  in \emph{Proceedings of International Conference on Knowledge Discovery and
  Data Mining}, vol.~96, no.~34, 1996, pp. 226--231.

\bibitem{campello2013density}
R.~J. Campello, D.~Moulavi, and J.~Sander, ``Density-based clustering based on
  hierarchical density estimates,'' in \emph{Proceedings of Pacific-Asia
  Conference on Knowledge Discovery and Data Mining}, 2013, pp. 160--172.

\bibitem{campello2015hierarchical}
R.~J. Campello, D.~Moulavi, A.~Zimek, and J.~Sander, ``Hierarchical density
  estimates for data clustering, visualization, and outlier detection,''
  \emph{ACM Transactions on Knowledge Discovery from Data}, vol.~10, no.~1, pp.
  1--51, 2015.

\bibitem{kodirov2016person}
E.~Kodirov, T.~Xiang, Z.~Fu, and S.~Gong, ``Person re-identification by
  unsupervised $\ell_1 $ graph learning,'' in \emph{Proceedings of European
  Conference on Computer Vision}, 2016, pp. 178--195.

\bibitem{khan2016unsupervised}
F.~M. Khan and F.~Bremond, ``Unsupervised data association for metric learning
  in the context of multi-shot person re-identification,'' in \emph{Proceedings
  of IEEE International Conference on Advanced Video and Signal Based
  Surveillance}, 2016, pp. 256--262.

\bibitem{li2019Unsupervised}
M.~Li, X.~Zhu, and S.~Gong, ``Unsupervised tracklet person re-identification,''
  \emph{IEEE Transactions on Pattern Analysis and Machine Intelligence},
  vol.~42, no.~7, pp. 1770--1782, 2019.

\bibitem{wu2019progressive}
Y.~Wu, Y.~Lin, X.~Dong, Y.~Yan, W.~Bian, and Y.~Yang, ``Progressive learning
  for person re-identification with one example,'' \emph{IEEE Transactions on
  Image Processing}, vol.~28, no.~6, pp. 2872--2881, 2019.

\bibitem{raychaudhuri2020exploiting}
D.~S. Raychaudhuri and A.~K. Roy-Chowdhury, ``Exploiting temporal coherence for
  self-supervised one-shot video re-identification,'' \emph{arXiv preprint
  arXiv:2007.11064}, 2020.

\end{thebibliography}

\begin{IEEEbiography}
[{\includegraphics[width=1in,height=1.25in,clip,keepaspectratio]{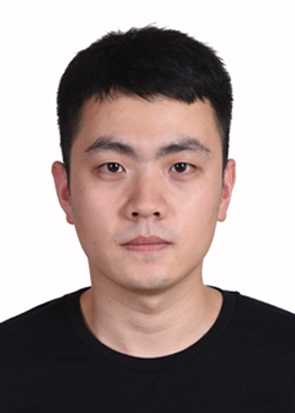}}]{Xueping Wang} 
is currently pursuing the Ph.D. degree with the College of Electrical and Information Engineering, Hunan University, China. His research interests include computer vision, person re-identification, and adversarial attack and defense methods.
\end{IEEEbiography}

\begin{IEEEbiography}
[{\includegraphics[width=1in,height=1.25in,clip,keepaspectratio]{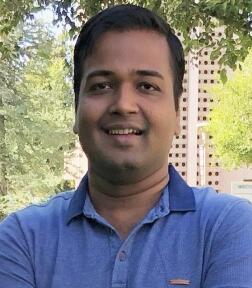}}]{Rameswar Panda} 
graduated from University of California, Riverside with a Ph.D. in Electrical and Computer Engineering in 2018. Previously, he received his Bachelors and Masters degree from Biju Patanaik University of Technology, India and Jadavpur University, India. He is currently a researcher at IBM Research AI (MIT-IBM Watson AI Lab). His main research interests include computer vision, machine learning, video summarization, person re-identification and multimedia. 
\end{IEEEbiography}


\begin{IEEEbiography}
[{\includegraphics[width=1in,height=1.25in,clip,keepaspectratio]{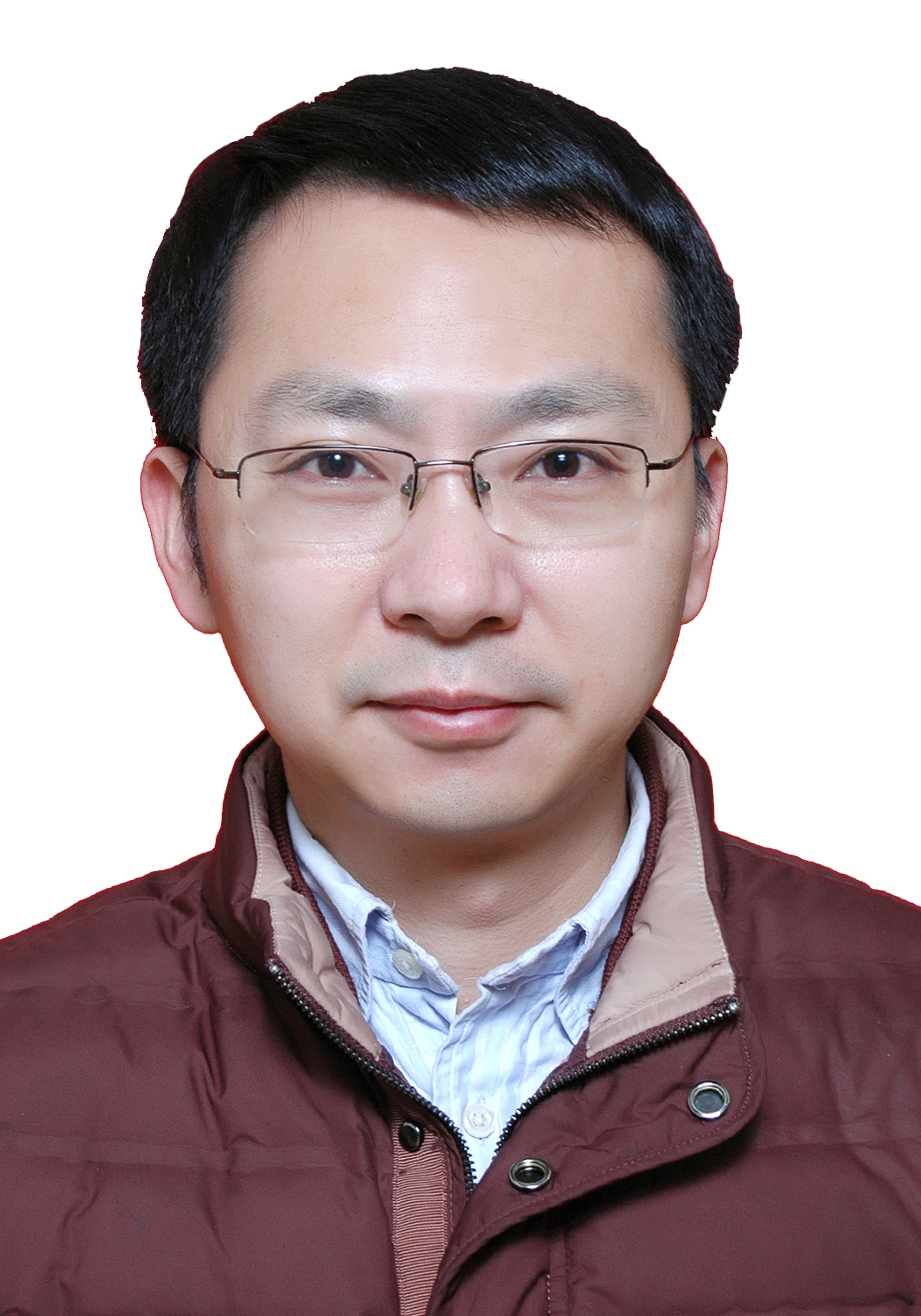}}]{Min Liu} 
is a professor at Hunan University. He received his bachelor degree from Peking University and Ph.D. degree in Electrical Engineering from the University of California, Riverside in 2012. He was a research intern in HHMI Janelia Farm Research Campus and a research scientist at the University of California, Santa Barbara. His research interests include computer vision and biomedical image analysis. Dr. Liu is an Associate Editor of BMC Bioinformatics.
\end{IEEEbiography}

\begin{IEEEbiography}
[{\includegraphics[width=1in,height=1.25in,clip,keepaspectratio]{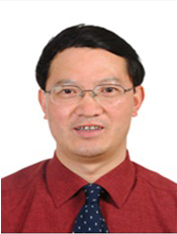}}]{Yaonan Wang} 
received the Ph.D. degree in electrical engineering from Hunan University, Changsha, China, in 1994. Since 1995, he has been a Professor with the College of Electrical and Information Engineering, Hunan University. From 1994 to 1995, he was a Post-Doctoral Research Fellow with the Normal University of Defense Technology, Changsha. From
1998 to 2000, he was supported as a Senior Humboldt Fellow by the Federal Republic of Germany at the University of Bremen, Bremen, Germany. From 2001 to 2004, he was a Visiting Professor at the University of Bremen. He is a member of the Chinese Academy of Engineering. His research interests include robotics and image processing.
\end{IEEEbiography}

\begin{IEEEbiography}
[{\includegraphics[width=1in,height=1.25in,clip,keepaspectratio]{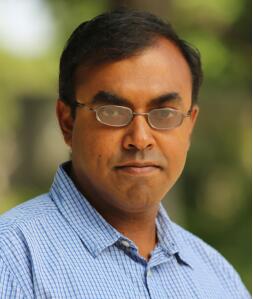}}]{Amit K. Roy-Chowdhury} 
received the Bachelors degree in Electrical Engineering from Jadavpur University, Calcutta, India, the Masters degree in Systems Science and Automation from the Indian Institute of Science, Bangalore, India, and the Ph.D. degree in Electrical and Computer Engineering from the University of Maryland, College Park. He is a Professor of Electrical and Computer Engineering and a Cooperating Faculty in the Department of Computer Science and Engineering, University of California, Riverside. His broad research interests include computer vision, image processing, and vision-based statistical learning, with applications in cyber-physical, autonomous and intelligent systems. He is a coauthor of two books: Camera Networks: The Acquisition and Analysis of Videos over Wide Areas, and Recognition of Humans and Their Activities Using Video. He is the editor of the book Distributed Video Sensor Networks. He has been on the organizing and program committees of multiple computer vision and image processing conferences and is serving on the editorial boards of multiple journals. He is a Fellow of the IEEE and IAPR.
\end{IEEEbiography}

\end{document}